\documentclass{article}

\usepackage[utf8]{inputenc}
\usepackage[T1]{fontenc}
\usepackage{lmodern}
\usepackage{amsmath}
\usepackage{amsfonts}
\usepackage{amssymb}
\usepackage{adjustbox}
\usepackage{multirow}
\usepackage{graphicx}
\usepackage{hyperref}
\usepackage[a4paper, total={6in, 8in}]{geometry}
\usepackage{dsfont}
\usepackage{booktabs}
\usepackage{bm}
\usepackage{subcaption}
\usepackage[ruled,vlined,linesnumbered]{algorithm2e}
\usepackage{authblk}

\hypersetup{
    colorlinks=true,
    linkcolor=blue,
    urlcolor=blue,
    citecolor=blue
}

\begin{document}

\title{Pi-Transformer: A Prior-Informed Dual-Attention Model for Multivariate Time-Series Anomaly Detection}

\author[1]{Sepehr Maleki\thanks{\href{mailto:smaleki@lincoln.ac.uk}{smaleki@lincoln.ac.uk}}}
\author[2]{Negar Pourmoazemi\thanks{\href{mailto:negar.pourmoazemi@thetrainline.com}{negar.pourmoazemi@thetrainline.com}}}
\affil[1]{Lincoln AI Lab, University of Lincoln, Lincoln, The UK}
\affil[2]{Trainline, London, The UK}

\date{}

\date{}

\maketitle
\begin{abstract}
\noindent Anomalies in multivariate time series often arise from temporal context and cross-channel coordination rather than isolated outliers. We present Pi-Transformer (Prior-Informed Transformer), a transformer with two attention pathways: data-driven series attention and a smoothly evolving prior attention that encodes temporal invariants such as scale-related self-similarity and phase synchrony. The prior provides an amplitude-insensitive temporal reference that calibrates reconstruction error. During training, we pair a reconstruction objective with a divergence term that encourages agreement between the two attentions while keeping them meaningfully distinct. The prior is regularised to evolve smoothly and is lightly distilled towards dataset-level statistics. At inference, the model combines an alignment-weighted reconstruction signal (Energy) with a mismatch signal that highlights timing and phase disruptions, and fuses them into a single score for detection. Across five benchmarks (SMD, MSL, SMAP, SWaT, and PSM), Pi-Transformer achieves state-of-the-art or highly competitive F1, with particular strength on timing and phase-breaking anomalies. Case analyses show complementary behaviour of the two streams and interpretable detections around regime changes. Embedding prior attention into transformer scoring yields a calibrated and robust approach to anomaly detection in complex multivariate systems.
\end{abstract}


\section{Introduction}
\label{sec:intro}

\noindent Anomaly detection in multivariate time series is a key part of modern monitoring systems. It emerges across important application areas such as industry \cite{maleki2016development, maleki2019robust, 10.1115/GT2018-75374, 7998334}, healthcare \cite{yang2023deep, khanizadeh2025smart}, cybersecurity \cite{landauer2025review, rashid2022anomaly}, finance \cite{song2023anomaly, li2021dynamic}, and space exploration \cite{lakhmiri2022anomaly, ratadiya2024raman}. In industrial plants, quickly finding sensor issues helps with predictive maintenance, which prevents expensive equipment failures and production downtime. Healthcare monitoring systems depend on anomaly detection to catch small changes in vital signs that might signal serious medical events. Infrastructure providers use anomaly detection to find intrusions, failures, or performance drops before they lead to widespread outages in telecommunications systems. Space missions rely on constant monitoring of spacecraft data, as missing anomalies can have severe consequences. The growing use of sensor-rich Internet of Things (IoT) and cyber-physical systems has greatly increased the scale, complexity, and detail of time series data. This makes effective anomaly detection both more important and more challenging.\medskip

\noindent \textbf{The challenge.} Detecting anomalies in multivariate time series can be challenging because anomalies do not usually appear as simple outliers. They often depend on context, involve multiple variables, and change over time. Anomalies only show up when we look at the timing and relationships between variables together. Real-world systems work across several timescales. Rapid changes, gradual shifts, daily or seasonal patterns, and long-term trends all exist and interact. Anomalies may affect one timescale while leaving others untouched. This means detection needs models that can understand multi-scale timing \cite{yang2023dcdetector}. The relationship between variables is also key. In many systems, the timing and synchronisation of sensor values matter more than the individual values themselves \cite{maleki2016development}. For instance, in rotating machinery, vibration signals from different components should stay in sync during normal operation. Small changes in this timing can signal mechanical failure, even if the vibrations remain within normal limits \cite{hu2022intelligent}. Because of these factors, it is not enough to depend on slight changes or simple distance measures. Additionally, operational data is often noisy and changes over time. Variance levels can shift between different states, and unusual but harmless setups are common. A method that mistakes this variability for anomalies will trigger false alarms, while one that overlooks it might miss real failures. Therefore, the main challenge is to create anomaly detection systems that are sensitive to small changes yet strong enough to handle normal variability, while also being suitable for large-scale use. In practice, several additional issues remain unresolved: (i) labelled anomalies are scarce and often incomplete, which limits supervised calibration and makes evaluation sensitive to assumptions; (ii) detectors must be robust to noise, missing or corrupted channels, and benign distribution shifts without producing alarm fatigue; and (iii) decisions must be interpretable so practitioners can validate alarms and diagnose root causes, especially when operating conditions evolve over time.
\medskip

\noindent \textbf{Existing solutions.} A variety of approaches have been used to tackle the anomaly detection challenge. Early methods relied on statistics and models, such as ARIMA for forecasting and EWMA for tracking changes from smoothed baselines \cite{box2015time, roberts2000control}. These methods are interpretable and efficient, but they struggle with non-linearities and multivariable dependencies. Traditional machine learning techniques like isolation forests and one-class SVMs \cite{scholkopf2001estimating} offered better generality but still weakly considered time correlations. This limitation reduced their effectiveness in changing environments. Deep learning methods transformed the field by allowing non-linear modelling of complex time series. Autoencoders and recurrent networks \cite{hinton2006reducing, graves2012long} learned normal patterns and identified anomalies based on reconstruction or prediction errors. However, these methods can be hard to calibrate because residual magnitudes drift with regime changes and variance, and thresholding is often heuristic \cite{blazquez2021review,hundman2018lstm}. Benign noise, rare but normal patterns, and shifts in data distribution often lead to inflated error metrics, resulting in false positives. Variational Autoencoders \cite{kingma2013auto, su2019robust} added probabilistic reconstruction, which made them more robust, but they also introduced training instability and increased computational demands. Recently, attention-based models like the Anomaly Transformer \cite{xu2021anomaly} and contrastive frameworks such as DCDetector~\cite{yang2023dcdetector} have achieved strong results on benchmark datasets. By using self-attention, these models better capture long-range relationships and interactions between multiple variables compared to earlier methods. Still, they have a major limitation: they do not have inductive biases to distinguish whether reconstruction or prediction errors indicate meaningful structural anomalies or just minor fluctuations. This shortcoming often results in missing subtle timing disruptions and generating false alarms in high-variance but normal situations.\medskip

\noindent \textbf{Our contribution.} We propose Pi-Transformer (Prior-Informed Transformer), a dual-attention architecture for multivariate time-series anomaly detection that mitigates calibration errors in reconstruction and prediction-based detectors under complex temporal context and cross-variable coordination. We use the term prior-informed to emphasise that the prior pathway is an inductive-bias reference over time indices (not a governing-equation constraint) designed to calibrate reconstruction evidence under changing regimes. Pi-Transformer couples data-driven series attention with a prior attention pathway that provides an index-based temporal reference that is insensitive to absolute magnitude. While prior--series attention was introduced by Anomaly Transformer~\cite{xu2021anomaly}, our method differs in three decisive ways that are essential for detection: \textbf{(i)} Prior construction: instead of a fixed, generic kernel prior, we construct a structured prior parameterised by time-varying scale and phase fields, capturing scale-related self-similarity and phase synchrony; this yields a stable reference under nominal regime changes while remaining sensitive to timing- and coordination-breaking anomalies. \textbf{(ii)} Learning strategy: we train the two pathways with a stop-gradient, alternating symmetric-divergence objective that aligns them without collapsing the mismatch signal, and we regularise the prior to evolve smoothly via explicit smoothness and weak distillation constraints. \textbf{(iii)} Inference mechanism: beyond using divergence only as a training regulariser, we explicitly decompose detection into two complementary signals (an alignment-weighted reconstruction (Energy) and an attention-mismatch score) and fuse them into the final anomaly score. Together, these choices turn the prior from a static inductive bias into a calibrated reference that improves robustness to contextual variability and strengthens detection of timing and coordination anomalies that are poorly captured by reconstruction error alone. We evaluate Pi-Transformer on five benchmarks: SMD~\cite{SMD}, MSL~\cite{MSL}, SMAP~\cite{MSL}, SWaT~\cite{SWaT}, and PSM~\cite{PSM}. The results show consistently strong performance, with particularly clear gains on timing and phase-breaking anomalies.\medskip

\noindent The remainder of this paper is organised as follows. Section \ref{sec:related} reviews related work in multivariate time-series anomaly detection. Section \ref{sec:methodology} presents Pi-Transformer, including prior attention construction, the stop-gradient alternating training strategy, and the inference-time fusion of Energy and mismatch signals. Section \ref{sec:experimental} reports experimental results, including comparisons, ablations, and sensitivity analyses. Section~\ref{sec:conclusion} concludes with limitations and future directions. \bigskip

\section{Related work}
\label{sec:related}

\noindent We review prior work in multivariate time-series anomaly detection with emphasis on attention-based detectors and calibration mechanisms that are most relevant to Pi-Transformer.\medskip

\noindent Statistical models were the first tools for detecting anomalies in temporal data. Autoregressive Integrated Moving Average (ARIMA) models combine autoregressive lags with differencing and moving averages to make forecasts \cite{box2015time}. Anomalies are identified when observations significantly differ from predictions. In a similar way, Exponentially Weighted Moving Average (EWMA) models \cite{roberts2000control} keep smoothed statistics and use fixed thresholds to detect anomalies. These methods are straightforward, easy to understand, and work well in stationary univariate situations like financial time series. However, their assumptions of linearity and stationarity do not hold in multivariate data. In such cases, temporal patterns can be nonlinear, noise levels can vary, and dependencies between sensors can offer important information.\medskip

\noindent Classical machine learning expanded anomaly detection beyond strict parametric assumptions. One-Class SVMs \cite{scholkopf2001estimating} learn separating hyperplanes in feature space to separate normal data from anomalies. Isolation Forests \cite{liu2008isolation} use random partitions to find samples that can be isolated in a few steps, marking them as likely anomalies. These models are robust against noise and have been used in manufacturing and network monitoring. However, they struggle with capturing sequential structure. Anomalies that occur in relation to temporal context or the synchrony of different variables often go undetected when samples are considered as independent points.\medskip

\noindent Deep learning methods have changed the field by focusing on representation learning. Reconstruction-based approaches dominated the first wave. Autoencoders \cite{hinton2006reducing} compress data windows into latent embeddings and then reconstruct them, identifying anomalies through high reconstruction error. Variational Autoencoders (VAEs) \cite{kingma2013auto} build on this framework with probabilistic modelling. They provide uncertainty estimates and likelihood-based scoring. OmniAnomaly \cite{su2019robust} combines a VAE with stochastic recurrent flows to capture temporal uncertainty and achieves strong results on spacecraft telemetry benchmarks like MSL and SMAP. However, reconstruction-based detectors are difficult to calibrate. Rare but normal patterns can inflate error, while anomalies that can be somewhat reconstructed may show low error, leading to false positives and false negatives.\medskip

\noindent A recurring limitation of reconstruction- and prediction-based detectors is that their residual scores are typically calibrated by post-hoc thresholding, yet the residual distribution is strongly regime- and variance-dependent: benign non-stationarity can inflate residuals, while expressive models may partially reconstruct anomalous segments, producing deceptively small errors. As a result, practical systems often rely on adaptive or dynamic thresholding or probabilistic scoring to track residual drift rather than a single fixed cutoff \cite{xu2021anomaly}. A complementary line of work targets residual instability by making normalisation and denoising learnable and robust to contaminated training windows. NormFAAE introduces a filter-augmented autoencoder with learnable normalization to reduce sensitivity to mixed regimes and noise when scoring anomalies \cite{yu2024normfaae}.\medskip

\noindent To mitigate these calibration problems, more recent work models normal behaviour more explicitly and depends less on raw reconstruction error. Likelihood-based scoring with VAEs assesses each window based on its reconstruction probability instead of point-wise error. This approach reduces alarms caused by variance \cite{an2015variational}. GAN-based models like MAD-GAN \cite{li2019mad} learn the regularities of multivariate temporal data through a generator-discriminator game and score anomalies using responses from the discriminator along with reconstruction discrepancies. Hybrid models combine the variational objective with adversarial training, such as VAEAT \cite{he2024vaeat}, to improve detector performance under complex normal conditions and changes in data distribution. Another approach replaces explicit reconstruction targets with masked prediction in a self-supervised manner. The MAD task \cite{fu2022mad} randomly masks segments and trains a model to recover them, resulting in embeddings and losses that highlight unusual structures without needing labels. These methods clarify decision boundaries when reconstruction signals are unclear and improve stability under non-stationarity.\medskip

\noindent Forecasting methods use prediction error as the signal for anomalies. Long Short-Term Memory (LSTM) networks \cite{graves2012long, SepehrLSTM} predict future values based on past history. They flag large deviations as anomalies. For example, DeepAR \cite{flunkert2017deepar} builds on this idea by producing full probabilistic forecasts. It detects anomalies when observations fall into low-likelihood areas. These methods work well when anomalies disrupt smooth dynamics, such as sudden changes in demand forecasting. However, they struggle with distribution shifts, needing re-training or recalibration when system behaviour changes. Additionally, forecasting models often mistake structural anomalies that remain predictable, such as phase-shifted oscillations, for normal behaviour. This happens because point-wise predictions may still remain accurate even with disruptions in underlying synchrony.\medskip

\noindent Attention-based methods represent a recent and significant development. TranAD \cite{TranAD} is one of the first detectors based on transformers. It combines a reconstruction goal with adversarial training. Its generator reconstructs windows, while a discriminator tells normal reconstructions apart from anomalous ones. This setup improves resistance to noise and unseen patterns. TranAD shows impressive results on large server datasets like SMD, highlighting the advantages of transformers in capturing long-range dependencies. However, it still faces the same calibration issues as earlier reconstruction-based models. In these cases, errors can reflect normal variability instead of true anomalies. Related transformer designs also explicitly separate temporal and inter-variable modeling. VTT factorises attention into temporal self-attention and variable self-attention to better capture cross-sensor dependencies and to support interpretation of anomalous variables \cite{kang2024vtt}.\medskip

\noindent The Anomaly Transformer \cite{xu2021anomaly} advances the field by directly modelling association discrepancies. It introduces a Gaussian-kernel attention mechanism that compares associations between normal and abnormal points. It also uses a MinMax learning objective that highlights differences. This design makes it very sensitive to small structural changes in temporal dependencies, allowing it to achieve state-of-the-art results on several benchmarks. However, its focus on association strength as the main signal makes it prone to noise. Associations may weaken in benign high-variance situations, leading to false alarms. At the same time, anomalies that maintain partial associations may produce weak signals. Additionally, without clear guidelines on temporal invariants like phase synchrony, the model struggles to differentiate between significant disruptions and harmless variability. Pi-Transformer adopts the dual-attention template of Anomaly Transformer~\cite{xu2021anomaly} but utilises novel prior parameterisation, training scheme, and inference-time use of mismatch.\medskip

\noindent DCDetector \cite{yang2023dcdetector} represents a new approach, moving from reconstruction to contrastive learning. It features a dual-attention design that includes patch-wise and in-patch augmentations. This trains the model to create consistent representations for various views of normal data. By avoiding reconstruction losses, DCDetector reduces the chance of anomalies affecting the training goal. It also performs well on datasets like MSL and SMAP, especially for anomalies with subtle local signs. However, its channel-independent patching limits its ability to capture synchrony across variables when anomalies show up as coordination problems. Additionally, contrastive learning makes it sensitive to augmentation choices and increases computational demands.\medskip

\noindent Building on these foundations, several transformer extensions focus on stability, coverage, and efficiency. TransAnomaly \cite{zhang2021unsupervised} combines long-range context with probabilistic reconstruction by linking a transformer encoder with a variational decoder. This approach improves calibration under uncertainty. Memory-guided designs like MEMTO \cite{song2023memto} keep standard prototypes in external memory. This helps reduce drift and strengthen recall during regime changes. Frequency- and channel-aware variants capture periodic links. For example, TimesNet \cite{wu2022timesnet} converts sequences into frequency-domain representations to model seasonality. Similarly, CATCH \cite{wu2024catch} uses frequency patching with channel awareness to highlight spectral anomalies. Patch-based alternatives substitute full self-attention with more efficient patch mixing. Decomposition-guided transformer designs further aim to stabilise representation learning under noise and complex seasonality. TransDe \cite{zhang2025transde} combines trend and seasonal decompositions with multi-scale patch transformers and a contrastive objective to improve robustness. PatchAD \cite{zhong2024patchad} employs MLP-Mixer blocks for windowed patch processing. Hybrid efficiency variants have also been explored by combining channel--time mixing with transformer-style dependency modelling. Mixer-Transformer \cite{fang2025mixertransformer} couples MLP-Mixer style mixing with anomaly-transformer-inspired association modelling and uses an adaptive threshold update to track distribution changes. SimAD \cite{zhong2025simad} demonstrates that simple dissimilarity scoring on patches can still perform well on standard benchmarks. Pre-training and sparsity have also been studied. AnomalyBERT \cite{jeong2023anomalybert} modifies bidirectional masked pre-training for transferable contextual embeddings. STformer \cite{li2024stformer} explicitly models joint spatial and temporal dependencies using multi-head patch attention across channels and time segments while stacking transformer encoder blocks for reconstruction-based detection. Reconstruction similarity hybrids like RESTAD \cite{ghorbani2024restad} cross-check evidence sources to mitigate calibration issues. Practical retraining methods such as LARA \cite{chen2024lara} offer light, anti-overfitting refresh processes that help maintain performance during domain shifts. Most of these designs improve representation learning, efficiency, or robustness, but they typically retain residual-driven or association-driven scoring without introducing an explicit temporal reference that is trained to remain stable under benign regime variation. In contrast, Pi-Transformer learns a structured prior attention that encodes temporal invariants and uses the resulting series--prior mismatch not only as a training regulariser but also as an explicit inference-time signal fused with alignment-weighted reconstruction evidence.\medskip

\noindent In parallel, graph-based methods encode the inter-channel structure that sequence-only models might miss. Graph attention over sensor nodes directly models pairwise relations, as in G-TSAD \cite{zhao2020multivariate}. GDN \cite{deng2021graph} learns dependency graphs from data to capture stable couplings in multivariate telemetry. Variational graph convolution combined with recurrence, as shown in DVGCRN \cite{chen2022deep}, supports this inductive bias by modelling uncertainty and temporal dynamics on the graph. Sparsity-promoting graph Autoencoders, like FuSAGNet \cite{han2022learning}, foster compact latent graphs that reflect persistent dependencies. By framing the issue as dynamic graph forecasting, DyGraphAD \cite{chen2302multivariate} shifts detection to deviations from predicted, time-varying relationships. At the subsequence level, Series2Graph \cite{boniol2022series2graph} converts time series into graph representations to enable density-based discovery of unusual patterns. These methods enhance transformer attention by introducing structural priors that stay stable during normal operation. These approaches primarily encode cross-channel structure, whereas Pi-Transformer targets temporal invariants within-window via a prior over time indices and uses attention disagreement to isolate coordination and timing disruptions during scoring.\medskip

\noindent Related ideas appear in other sensing domains where anomaly detection is framed as deviation from an explicit, interpretable prior model rather than purely from data-fit error. For example, in hyperspectral anomaly detection, model-driven priors are often combined with deep networks via interpretable deep unfolding or data-driven paradigms \cite{li2023lrrnet,li2024ldp}, and transformer pretraining has been explored to encode modality-specific structure at scale \cite{hong2024spectralgpt}. Although these methods are not direct baselines for multivariate time series, they motivate the same design choice adopted here. \medskip

\noindent Existing detectors often rely on residuals or association discrepancies that remain sensitive to benign non-stationarity and do not explicitly encode temporal invariants (e.g., phase consistency) that can stabilise scoring. These limitations motivate the prior-informed dual-attention design introduced next.\bigskip

\section{Methodology}
\label{sec:methodology}

\subsection{Problem Formulation}
\label{sec:problemformulation}

\noindent We consider the problem of unsupervised anomaly detection in multivariate time series. Here, anomalies are defined as deviations from learned normal behaviour rather than being explicitly labeled examples. Let 

\[
\mathbf{X}=\{\mathbf{x}_1,\ldots,\mathbf{x}_T\}, 
\qquad \mathbf{x}_t \in \mathbb{R}^C,
\]

\noindent denote a multivariate sequence of length $T$ with $C$ channels (sensors). In many real-world situations, such as monitoring industrial equipment, large amounts of data are available from normal operation. However, failures are rare or absent. This limitation calls for an approach that learns the structure of normality without supervision and identifies deviations as potential anomalies.

\noindent A major challenge in anomaly detection is that anomalies in time series often depend on the temporal context and the coordination between channels. A reading may seem normal when viewed alone but could be anomalous relative to its recent history (contextual anomalies). Additionally, multiple sensors may each stay within normal ranges, but their phase relationships (the timing and synchronisation between channels) can reveal a fault. For instance, in rotating machinery, vibration sensors at different locations should maintain consistent phase offsets due to mechanical coupling \cite{hu2022intelligent}. A bearing fault may keep the amplitudes the same but distort these offsets \cite{stack2004amplitude}. In thermodynamic systems, the relationship between temperature and pressure sensors indicates efficiency. Disruptions in their synchronisation can signal degradation even before individual measurements exceed thresholds \cite{kaushik2024detecting}.\medskip

\noindent We therefore frame the unsupervised anomaly detection problem as modelling the nominal window distribution with a reconstruction model that couples to an explicit prior over temporal relationships. Given a dataset of windows taken from normal operation:

\[
\mathcal{D}_{\text{train}}=\{\mathbf{W}_i\in\mathbb{R}^{L\times C}\}_{i=1}^N,
\]

\noindent Our goal is to learn parameters $\theta$ so the model can reconstruct each window $\mathbf{W}_i$ while also capturing two key properties of nominal sequences: (1) self-similarity across timescales, often defined by Hurst-related features, and (2) stable cross-channel phase relationships, which reflect coordinated system dynamics.\medskip

\noindent During inference, the model evaluates a test window $\mathbf{W}$ against these learned patterns. Windows that cannot be reconstructed consistently with their expected self-similarity and phase structure receive high anomaly scores and are marked as anomalous. This approach changes anomaly detection from simple classification to discovering and monitoring the temporal patterns that define normal system behaviour.

\subsection{Pi-Transformer}
\label{sec:PiTrans}

\begin{figure}[!htb]
	\centering
	\includegraphics[trim={3cm 0cm .8cm 0cm},clip, width=\textwidth]{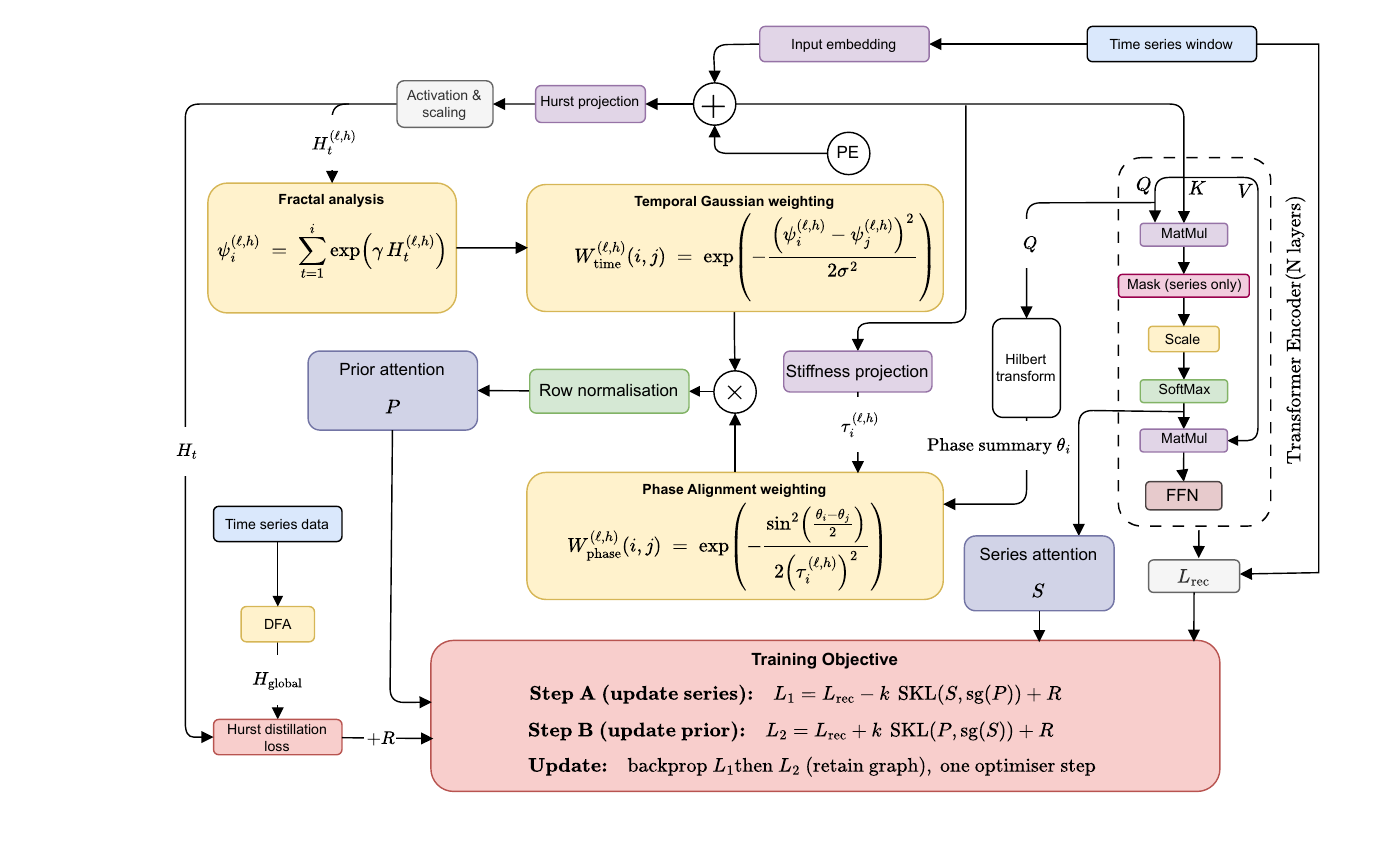}
	\caption{The Pi-Transformer architecture.}
	\label{fig:modelarc}
\end{figure}

\noindent Pi-Transformer (Figure~\ref{fig:modelarc}) uses a dual-pathway attention mechanism comprising a data-driven series attention learned from the data and a prior-informed attention that encodes expected timing relationships under nominal conditions. Their agreement within each window is transformed, through a time-wise softmax, into prior-alignment weights that sum to one. Indices that remain prior-consistent with the prior receive larger weights, while poorly aligned indices receive smaller weights. When the series and prior attentions are well aligned, the weights tend to be less peaked, whereas around disruptions they concentrate on the most prior-consistent indices. The resulting time-wise anomaly score, which we refer to as the Energy signal, is defined as
\begin{equation}
e_i \;=\; w_i \, r_i,
\label{eq:energy}
\end{equation}
\noindent where $r_i$ is the reconstruction error at time index $i$ and $w_i$ is a within-window prior-alignment weight. We define
\[
r_i=\frac{1}{C}\sum_{c=1}^{C}(x_{i,c}-\widehat{x}_{i,c})^2\;,
\]
\noindent where $x_i\in\mathbb{R}^C$ is the multivariate observation at time $i$ and $\widehat{x}_i$ is its reconstruction. The weight vector $w\in\mathbb{R}^L$ is obtained by transforming the series--prior mismatch within the same window into a probability distribution. Indices that are well-aligned with the prior receive larger weights, while poorly aligned indices are down-weighted. This converts raw reconstruction error into a time-wise Energy signal that is less sensitive to harmless regime variability.\medskip

\noindent Pi-Transformer consists of a prior-informed encoder with stacked Transformer layers and a linear reconstruction head. For each input window $X\in\mathbb{R}^{L\times C}$, the encoder produces two attention maps per layer and head, namely a data-driven series attention and a structured prior attention. The reconstruction head maps the resulting context representations back to the input space.

\medskip

\noindent We index window positions by $i,j,t \in \{1,\ldots,L\}$ and attention heads by $h \in \{1,\ldots,H\}$. The series attention follows the standard scaled dot-product formulation with a softmax over time,
\[
S^{(\ell,h)}=\operatorname{softmax}\!\left(\frac{Q^{(\ell,h)}K^{(\ell,h)\top}}{\sqrt{d_h}}\right),
\]

\noindent where $Q^{(\ell,h)}$ and $K^{(\ell,h)}$ are learned projections of the encoder features and $d_h$ is the head dimension. The softmax is applied row-wise, so row $i$ corresponds to the query, or current time step being updated, while column $j$ corresponds to a candidate key/value index within the same window. We apply a causal mask so that each row assigns probability mass only to indices $j \le i$. Accordingly, the row distribution $S^{(\ell,h)}_{i,\cdot}$ specifies how strongly time step $i$ attends to each admissible index $j$ when estimating $x_i$.

\noindent The prior attention specifies which indices are expected to contribute to each position under nominal timing and coordination. It uses the same row--column convention as $S^{(\ell,h)}$: row $i$ denotes the current query position, and column $j$ denotes a candidate source index in the window, again restricted to $j \le i$ by the causal mask. Unlike $S^{(\ell,h)}$, it is not derived from content similarity in the $QK^\top$ space. Instead, it is constructed from low-dimensional, time-indexed fields predicted by the encoder. Concretely, for each layer $\ell$ and head $h$, the encoder predicts a Hurst-like scale field $H^{(\ell,h)}_t$ and a phase-stiffness field $\tau^{(\ell,h)}_t$ for $t\in\{1,\dots,L\}$. The remaining components are computed deterministically from these fields. The scale field $H^{(\ell,h)}_t$ induces a warped time coordinate $\psi^{(\ell,h)}$ used by the temporal kernel $W^{(\ell,h)}_{\mathrm{time}}$, while $\tau^{(\ell,h)}_t$ controls the phase-consistency gate $W^{(\ell,h)}_{\mathrm{phase}}$. We construct these components step by step and then combine and row-normalise them to obtain the prior attention map $P^{(\ell,h)}$. We first form a monotone ``fractal abscissa'' that warps time according to the predicted scale field,

\[
\psi^{(\ell,h)}_i \;=\; \sum_{t=1}^{i}\exp\!\big(\gamma\,H^{(\ell,h)}_t\big).
\]

\noindent Here $\gamma>0$ controls the strength of the time warp induced by the scale field, and $\sigma>0$ sets the bandwidth of the Gaussian kernel in the warped coordinate. We define a temporal kernel over index pairs $(i,j)$ using a Gaussian in the warped coordinate,
\[
W^{(\ell,h)}_{\mathrm{time}}(i,j) \;=\; \exp\!\left(-\frac{\big(\psi^{(\ell,h)}_i-\psi^{(\ell,h)}_j\big)^2}{2\sigma^2}\right)\,\mathbb{I}[j\le i].
\]

\noindent We then compute a phase-consistency gate using a Hilbert-transform-based phase summary and a stiffness-controlled circular distance. We apply a Hilbert transform to a scalar projection of the encoder features to form an analytic signal $a^{(\ell,h)}_i = u^{(\ell,h)}_i + \mathrm{i}\,\mathcal{H}(u^{(\ell,h)})_i$, and define the instantaneous phase summary as $\theta^{(\ell,h)}_i = \arg\!\big(a^{(\ell,h)}_i\big)$. This yields an amplitude-insensitive phase descriptor that is used only to gate index-to-index consistency in the prior.

\[
W^{(\ell,h)}_{\mathrm{phase}}(i,j) \;=\; \exp\!\left(-\frac{\sin^2\!\left(\frac{\theta^{(\ell,h)}_i-\theta^{(\ell,h)}_j}{2}\right)}{2\big(\tau^{(\ell,h)}_i\big)^2}\right)\,\mathbb{I}[j\le i],
\]

\noindent where $\tau^{(\ell,h)}_i$ denotes the phase-stiffness at index $i$ in layer $\ell$ and head $h$. The unnormalised prior score is given by $A^{(\ell,h)}(i,j)=W^{(\ell,h)}_{\mathrm{time}}(i,j)\,W^{(\ell,h)}_{\mathrm{phase}}(i,j)$, and the row-stochastic prior attention is obtained by row normalisation,
\[
P^{(\ell,h)}_{i,j} \;=\; \frac{A^{(\ell,h)}(i,j)}{\sum_{m=1}^{L}A^{(\ell,h)}(i,m)}.
\]

\noindent We regularise the phase-stiffness field $\tau^{(\ell,h)}_t$ to evolve smoothly over $t$, and apply weak distillation of the scale field $H^{(\ell,h)}_t$ towards dataset-level nominal statistics. Mismatch between $S^{(\ell,h)}$ and $P^{(\ell,h)}$ is quantified using a symmetric KL divergence, and is used both as a training regulariser and as an explicit mismatch signal at inference.

\subsection{Training}
\label{sec:training}

\noindent Training the Pi-Transformer has two main goals. First, it aims to fit the nominal dynamics accurately enough to provide a calibrated reconstruction signal. Second, it seeks to learn a prior-informed attention that is stable and distinctly different from the data-driven series attention. The training objective combines the reconstruction quality with regularisation terms that stabilise the prior attention and keep a measurable difference from the series pathway. For simplicity, we denote the series and prior attentions as $S$ and $P$ respectively. These refer to their row-normalised distributions, which are the attention weights over window indices for each position $i$. We apply the same causal mask in both pathways so that each row $S_i$ and $P_i$ assigns probability mass only to indices $j\le i$, and the KL divergences are computed over the same support. These will later appear with specific layer and head indices in Eq. \eqref{eq:Delta}, but we are using shorthand here. The local difference between the two pathways at time index $i$ is measured by a symmetric Kullback–Leibler divergence:\medskip

\[
\delta_i \;=\; \mathrm{KL}(S_i \,\Vert\, P_i) \;+\; \mathrm{KL}(P_i \,\Vert\, S_i).
\]

\medskip
\noindent During training, we compute this divergence row-wise for each layer and head and then average it
across layers, heads, and time indices to obtain a scalar divergence term for the batch. We do not apply inverse-temperature scaling during training. This retains fine-grained mismatch structure while providing a stable optimisation objective for the stop-gradient alternation. During inference, we gather the same divergences across layers and heads and scale them by an inverse-temperature factor (see Eq.~\eqref{eq:Delta}). If we drove this divergence to zero directly, one pathway could collapse onto the other. We therefore apply stop-gradient alternation only to the divergence terms. In the first pass, the symmetric divergence is computed with $\mathrm{sg}(P)$, so the KL gradients flow only through the series pathway. In the second pass, the divergence is computed with $\mathrm{sg}(S)$, so the KL gradients flow only through the prior pathway. Importantly, the reconstruction loss $\mathcal{L}_{\mathrm{rec}}$ (and the prior regulariser $\mathcal{L}_{\mathrm{reg}}$) are applied in both passes, updating the shared encoder and reconstruction head and stabilising the prior fields, while the stop-gradient prevents the divergence term from causing mutual collapse. Formally, the two per-batch losses are:

\begin{align}
\mathcal{L}_1 &= \mathcal{L}_{\mathrm{rec}} \;+\; k\,
\Big[
\mathrm{KL}\!\big(S \,\Vert\, \mathrm{sg}(P)\big)
+
\mathrm{KL}\!\big(\mathrm{sg}(P) \,\Vert\, S\big)
\Big]
\;+\;
\lambda_{\mathrm{reg}}\mathcal{L}_{\mathrm{reg}}, \\[4pt]
\mathcal{L}_2 &= \mathcal{L}_{\mathrm{rec}} \;+\; k\,
\Big[
\mathrm{KL}\!\big(P \,\Vert\, \mathrm{sg}(S)\big)
+
\mathrm{KL}\!\big(\mathrm{sg}(S) \,\Vert\, P\big)
\Big]
\;+\;
\lambda_{\mathrm{reg}}\mathcal{L}_{\mathrm{reg}}.
\end{align}

\noindent Here $\mathrm{sg}(\cdot)$ denotes the stop-gradient operator, $\mathcal{L}_{\mathrm{reg}}$ collects regularisation terms applied to the prior, and $\lambda_{\mathrm{reg}}$ is its weight. We backpropagate both losses sequentially in each iteration. $\mathcal{L}_{\mathrm{rec}}$ updates the shared encoder and reconstruction head in both passes, while the stop-gradient ensures that the KL term in $\mathcal{L}_1$ backpropagates only through the series pathway and the KL term in $\mathcal{L}_2$ backpropagates only through the prior pathway.

\noindent The reconstruction objective provides the main data-fidelity signal and sets the scale of test-time errors. For each window of length $L$ with $C$ channels, we minimise the mean squared error averaged over time and channels:

\begin{equation}
\mathcal{L}_{\mathrm{rec}}
\;=\;
\frac{1}{L}\sum_{i=1}^{L}\frac{1}{C}\sum_{c=1}^{C}
\big(x_{i,c}-\widehat{x}_{i,c}\big)^2 ,
\label{eq:recon_train}
\end{equation}

\noindent where $x_{i,c}$ and $\widehat{x}_{i,c}$ are the observed and reconstructed values for channel $c$ at time $i$. This encourages the encoder to capture common nominal structure rather than memorising isolated fluctuations, yielding a calibrated baseline for time-wise errors.\medskip

\noindent To stabilise the prior-informed pathway, we regularise the phase-stiffness field for smooth evolution and distil the scale field towards nominal statistics. We penalise the prior stiffness field $\tau^{(\ell,h)}_i$ for rapid changes with a first-difference term:

\begin{equation}
\mathcal{R}_{\mathrm{smooth}}
\;=\;
\frac{1}{L_{\text{att}}H}\sum_{\ell=1}^{L_{\text{att}}}\sum_{h=1}^{H}
\frac{1}{L-1}\sum_{i=2}^{L}
\big(\tau^{(\ell,h)}_i-\tau^{(\ell,h)}_{i-1}\big)^2 .
\label{eq:smooth}
\end{equation}

\noindent Additional stabilisers are applied to the unnormalised prior scores (the pre-normalisation values that generate $P$), together with the smoothness penalty $\mathcal{R}_{\mathrm{smooth}}$ on $\tau^{(\ell,h)}_i$ and a weak distillation term that nudges the predicted Hurst field towards dataset-level statistics estimated from nominal data. We group these terms as

\[
\mathcal{L}_{\mathrm{reg}} \;=\;
\lambda_{\mathrm{smooth}}\mathcal{R}_{\mathrm{smooth}} \;+\;
\mathcal{R}_{\mathrm{distill}} \;+\;
\mathcal{R}_{\mathrm{prior}}.
\]

\noindent where $\mathcal{R}_{\mathrm{distill}}$ denotes the Hurst distillation penalty and $\mathcal{R}_{\mathrm{prior}}$ collects the remaining prior stabilisers applied to the pre-normalisation prior scores.

\noindent We use Adam optimiser, apply gradient clipping, and use early stopping based on validation reconstruction loss. Hyperparameters (learning rate, batch size, window length, and regularisation weights) are tuned on the validation split. This regime fits nominal data, stabilises the prior-informed attention, and preserves a controlled series–prior separation, which supports inference-time fusion of alignment-weighted reconstruction evidence with the mismatch signal (Algorithm~\ref{alg:training}).

\begin{algorithm}[!h]
\caption{Training Pi-Transformer}
\label{alg:training}
\KwIn{Training multivariate series $\mathcal{D}_{\text{train}}$, window length $L$, stride $s{=}1$, attention layers $L_{\text{att}}$, heads $H$, weights $\lambda_{\text{reg}}, k$.}
\KwOut{Trained Pi-Transformer.}

\While{not converged}{
  Sample minibatch of windows $\{\mathbf{W}\}\subset \mathcal{D}_{\text{train}}$\;
  Forward pass to obtain reconstructions $\hat X$, series attentions $\{S^{(\ell,h)}\}$, and prior attentions $\{P^{(\ell,h)}\}$\;

  Compute reconstruction loss
  $\mathcal{L}_{\mathrm{rec}}=\tfrac{1}{LC}\sum_{i,c}(x_{i,c}-\hat x_{i,c})^2$\;

  Compute regulariser $\mathcal{L}_{\mathrm{reg}}=\lambda_{\mathrm{smooth}}\mathcal{R}_{\mathrm{smooth}}+\mathcal{R}_{\mathrm{distill}}+\mathcal{R}_{\mathrm{prior}}$\;

  \BlankLine
  \textbf{Pass 1 (update series via divergence; stop-gradient on prior in divergence term):}\;
  Compute symmetric KL with stop-gradient on the \textbf{prior}:
  \[
  \mathcal{D}_{S}=\tfrac{1}{L_{\mathrm{att}}HL}\sum_{\ell,h}\sum_{i}\Big[
  \mathrm{KL}\!\big(S^{(\ell,h)}_{i,\cdot}\,\|\,\mathrm{sg}(P^{(\ell,h)}_{i,\cdot})\big)+
  \mathrm{KL}\!\big(\mathrm{sg}(P^{(\ell,h)}_{i,\cdot})\,\|\,S^{(\ell,h)}_{i,\cdot}\big)\Big]
  \]
  Form loss $\mathcal{L}_1=\mathcal{L}_{\mathrm{rec}}+k\mathcal{D}_{S}+\lambda_{\mathrm{reg}}\mathcal{L}_{\mathrm{reg}}$\;
  Zero gradients\;
  Backpropagate $\mathcal{L}_1$\;
  Optimiser step\;

  \BlankLine
  \textbf{Pass 2 (update prior via divergence; stop-gradient on series in divergence term):}\;
  Forward pass again to obtain $\hat X$, $\{S^{(\ell,h)}\}$, $\{P^{(\ell,h)}\}$\;
  Recompute $\mathcal{L}_{\mathrm{rec}}$ and $\mathcal{L}_{\mathrm{reg}}$\;
  Compute symmetric KL with stop-gradient on the \textbf{series}:
  \[
  \mathcal{D}_{P}=\tfrac{1}{L_{\mathrm{att}}HL}\sum_{\ell,h}\sum_{i}\Big[
  \mathrm{KL}\!\big(P^{(\ell,h)}_{i,\cdot}\,\|\,\mathrm{sg}(S^{(\ell,h)}_{i,\cdot})\big)+
  \mathrm{KL}\!\big(\mathrm{sg}(S^{(\ell,h)}_{i,\cdot})\,\|\,P^{(\ell,h)}_{i,\cdot}\big)\Big]
  \]
  Form loss $\mathcal{L}_2=\mathcal{L}_{\mathrm{rec}}+k\mathcal{D}_{P}+\lambda_{\mathrm{reg}}\mathcal{L}_{\mathrm{reg}}$\;
  Zero gradients\;
  Backpropagate $\mathcal{L}_2$\;
  Optimiser step\;
}
\end{algorithm}

\subsection{Inference}
\label{sec:inference}

\noindent At inference, the Pi-Transformer generates two time-wise quantities for each window position $i\in\{1,\dots,L\}$: (1) a reconstruction error $r_i$, defined as the channel-averaged squared error at time $i$, and (2) a series--prior mismatch $\Delta_i$ that uses the same symmetric divergence applied during training, but is now averaged across layers and heads and adjusted by an inverse-temperature. Specifically,

\begin{equation}
\Delta_i
\;=\;
\mathcal{T} \cdot \frac{1}{L_{\text{att}} H}
\sum_{\ell=1}^{L_{\text{att}}}
\sum_{h=1}^{H}
\Big[
\mathrm{KL}\!\big(S^{(\ell,h)}_{i,\cdot}\,\Vert\,P^{(\ell,h)}_{i,\cdot}\big)
+
\mathrm{KL}\!\big(P^{(\ell,h)}_{i,\cdot}\,\Vert\,S^{(\ell,h)}_{i,\cdot}\big)
\Big]\;,
\label{eq:Delta}
\end{equation}

\noindent where $S^{(\ell,h)}_{i,\cdot}$ and $P^{(\ell,h)}_{i,\cdot}$ denote the row distributions of the series and prior attentions at time $i$ in layer $\ell$ and head $h$, with probability mass restricted to valid indices $j\le i$ by the shared causal mask. This generalises the row-wise divergence $\delta_i$ of the training loss by averaging across layers and heads and introducing a scaling factor $\mathcal{T}>0$ that controls the sharpness of the softmax in Eq.~\eqref{eq:softmax_w}. The divergences $\{\Delta_j\}_{j=1}^L$ within each window are converted into prior-alignment weights $w$ via a softmax over the negatives:

\begin{equation}
w_i
\;=\;
\frac{\exp\!\big(-\Delta_i\big)}{\sum_{j=1}^{L}\exp\!\big(-\Delta_j\big)}\,,
\label{eq:softmax_w}
\end{equation}

\noindent which assigns larger weights to indices with stronger series–prior agreement. The Energy signal then follows Eq. \eqref{eq:energy}, so reconstruction error is emphasised at phase-consistent positions and suppressed where alignment collapses. Because $w$ is normalised within each window, emphasis is redistributed rather than scaled uniformly.\medskip

\noindent For detection, we combine the two complementary streams (Energy $e_i$ and mismatch $\Delta_i$) by robustly normalising each signal on the training split and then taking their point-wise maximum, yielding $\widetilde e_i$ and $\widetilde d_i$ as the normalised Energy and mismatch signals:

\begin{align}
\widetilde e_i = \max\!\left(0,\ \frac{e_i - \operatorname{med}_{\text{train}}(e)}{\operatorname{IQR}_{\text{train}}(e)}\right), \quad \quad
\widetilde d_i = \max\!\left(0,\ \frac{\Delta_i - \operatorname{med}_{\text{train}}(\Delta)}{\operatorname{IQR}_{\text{train}}(\Delta)}\right),
\end{align}
and then fused into a single score
\begin{equation}
f_i \;=\; \max\!\big(\widetilde e_i,\ \widetilde d_i\big).
\label{eq:fused}
\end{equation}

\noindent This max fusion triggers when either stream provides strong evidence, while keeping the final score in the same robustly normalised scale. A percentile threshold is chosen from fused scores on the training and threshold splits:

\begin{equation}
\eta_{\mathrm{thr}} \;=\; \mathrm{Percentile}_{\,100-\rho}\!\left(\{f_i\}_{\text{train}} \cup \{f_i\}_{\text{threshold}}\right),
\qquad
\widehat{y}_i \;=\; \mathbb{I}\{ f_i > \eta_{\mathrm{thr}} \},
\label{eq:thresh_fused}
\end{equation}

\noindent with target anomaly percentage $\rho \in (0,100)$. For evaluation, we apply the point-adjust procedure \cite{liu2023smoothing} to expand detections across labelled anomalous segments. This fused scoring rule ensures that amplitude or shape anomalies surface through elevated Energy, timing anomalies through elevated series-prior mismatch, and mixed anomalies through joint elevation of both streams. The inverse-temperature $\mathcal{T}$ further controls sharpness: larger $\mathcal{T}$ concentrates $w$ on the most series-prior consistent indices, which can shift Energy peaks slightly before or after an onset in overlapping windows.\medskip

\begin{figure}[!htb]
    \centering
    \includegraphics[width=\textwidth]{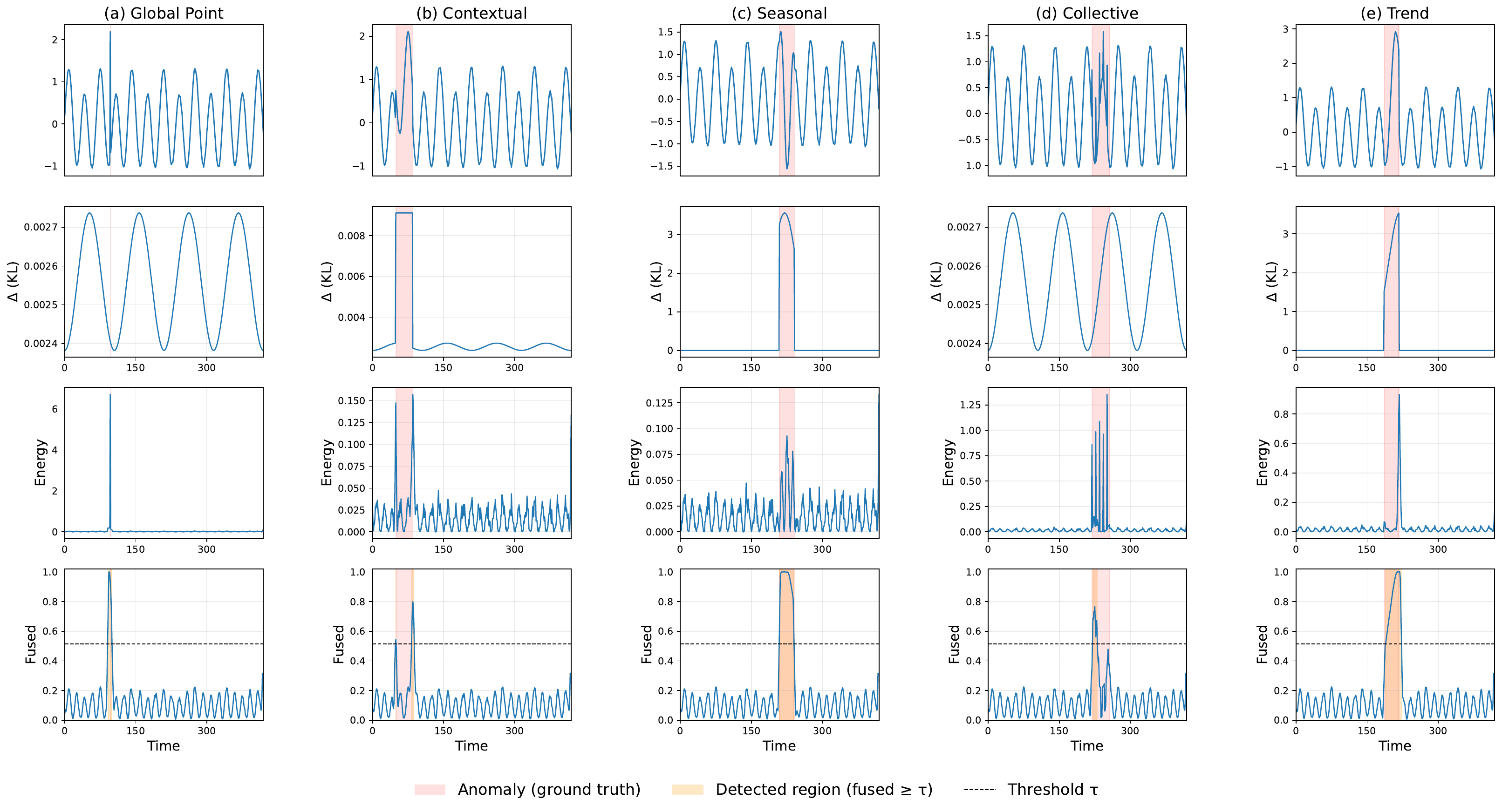}
    \caption{\textbf{Canonical anomaly types.}
    Each column shows (top to bottom): observed series with ground-truth anomalies (red), series--prior mismatch $\Delta_i$ (raw), Energy $e_i=w_ir_i$ (raw), and the fused score $f_i=\max(\widetilde e_i,\widetilde d_i)$ (unit-scaled). 
    A single per-dataset threshold $\eta_{\mathrm{thr}}$ (dashed) is applied across all types. Orange shading marks detected regions where $f_i>\eta_{\mathrm{thr}}$. Amplitude or shape anomalies (point, contextual, collective) drive Energy spikes with low $\Delta$, while timing and phase anomalies (seasonal, trend) elevate $\Delta$ at breakpoints even when Energy is muted.}
    \label{fig:dual_channels}
\end{figure}

\noindent Figure~\ref{fig:dual_channels} illustrates how the Pi-Transformer reacts to different types of anomalies. Point, contextual, and collective anomalies show sharp increases in the Energy signal $e_i=w_ir_i$ because the series pathway has difficulty reconstructing them, while the mismatch $\Delta_i$ stays low. On the other hand, seasonal and trend anomalies disrupt the timing, causing the mismatch $\Delta_i$ to increase significantly at the breakpoint, even when $e_i$ is low. In these examples, the combined score $f_i=\max(\widetilde e_i,\widetilde d_i)$ exceeds the threshold across all five types. This means the Pi-Transformer detects both differences in amplitude and shape (through $e_i$) as well as timing and phase changes (through $\Delta_i$) in a cohesive way.

\begin{figure}[!h]
  \centering
  \includegraphics[width=\textwidth]{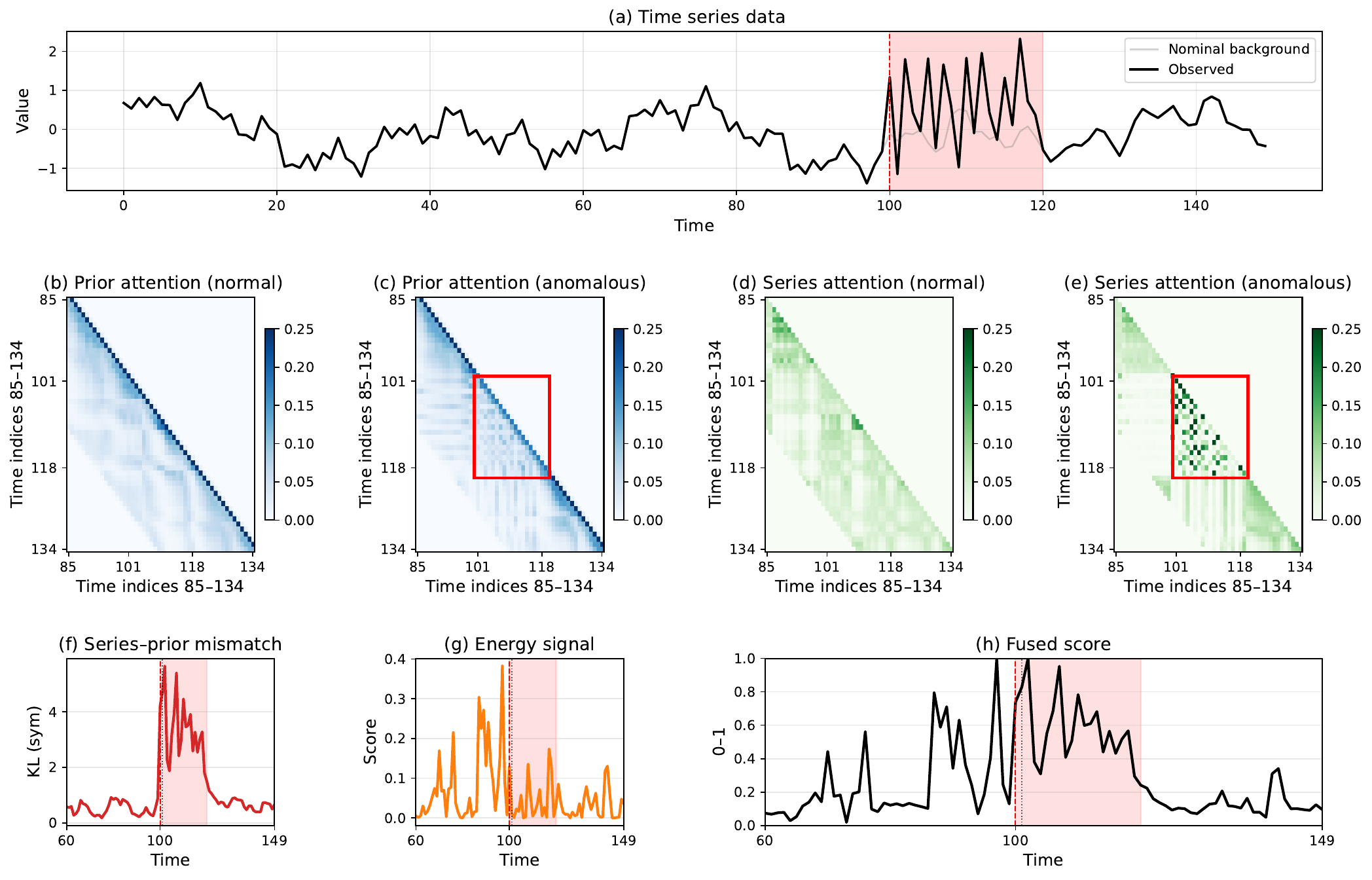}
  \caption{\textbf{Illustrative mechanism around a phase-breaking anomaly.}
  (a) Time series with anomalous interval (red). (b--e) Prior vs.\ series attentions in nominal and anomalous regimes (zoomed). 
  (f) Series--prior mismatch $\Delta_i$, (g) Energy $e_i=w_ir_i$, and (h) fused score $f_i=\max(\widetilde e_i,\widetilde d_i)$. 
  $\Delta$ spikes at the onset (alignment collapse), while Energy peaks immediately before or after where error is high but alignment is non-zero. The fused score rises as the anomaly enters the rolling window, providing robust detection.}
  \label{fig:pitransattention}
\end{figure}

\noindent Figure~\ref{fig:pitransattention} highlights a phase-breaking anomaly to show how the detection takes place. Here, the prior attention evolves smoothly, while the series attention reconfigures abruptly, producing a spike in mismatch $\Delta_i$. Alignment weights $w_i$ shift towards neighbouring series-prior consistent indices, so Energy $e_i=w_ir_i$ rises immediately before the onset, where reconstruction error is high but alignment is not fully collapsed. At the exact breakpoint, Energy is often muted but the mismatch channel is active. As a result, the fused score climbs as soon as the anomaly enters the rolling window, ensuring timely and reliable detection. The full sequence of steps is summarised in Algorithm \ref{alg:inference}.\medskip

\begin{algorithm}[!h]
\caption{Inference and anomaly scoring}
\label{alg:inference}
\KwIn{Trained Pi-Transformer, test series $X_{\text{test}}$, window length $L$, stride $s{=}1$, inverse-temperature $\mathcal{T}$, anomaly percentage $\rho$, training (and optional threshold) splits.}
\KwOut{Anomaly labels $\hat y$, fused scores $f$.}

\For{each window of $X_{\text{test}}$}{
  Compute reconstruction error $r_i$ (MSE across channels)\;
  Obtain series and prior attentions $S^{(\ell,h)}, P^{(\ell,h)}$\;
  Compute mismatch $\Delta_i = \mathcal{T}\cdot \tfrac{1}{L_{\mathrm{att}}H}\sum_{\ell,h}\big[\mathrm{KL}(S^{(\ell,h)}_{i,\cdot}\Vert P^{(\ell,h)}_{i,\cdot})+\mathrm{KL}(P^{(\ell,h)}_{i,\cdot}\Vert S^{(\ell,h)}_{i,\cdot})\big]$\;
  Convert to prior-alignment weights $w_i = \exp(-\Delta_i)/\sum_j \exp(-\Delta_j)$\;
  Compute Energy $e_i = w_i r_i$\;
}
Map window-level $r_i$, $\Delta_i$, and $e_i$ to the timeline by end-alignment and averaging contributions when multiple windows cover the same time index\;
Robustly normalise:
$\widetilde e_i=\max(0,\tfrac{e_i-\operatorname{med}_{\text{train}}(e)}{\operatorname{IQR}_{\text{train}}(e)}),\quad
\widetilde d_i=\max(0,\tfrac{\Delta_i-\operatorname{med}_{\text{train}}(\Delta)}{\operatorname{IQR}_{\text{train}}(\Delta)})$\;
Fuse: $f_i=\max(\widetilde e_i,\widetilde d_i)$\;
Global threshold:
$\eta_{\mathrm{thr}}=\mathrm{Percentile}_{100-\rho}(\{f_i\}_{\text{train}}\cup\{f_i\}_{\text{threshold}})$\;
Label anomalies: $\hat y_i=\mathbb{I}\{f_i>\eta_{\mathrm{thr}}\}$\;
Apply point-adjust for segment-level evaluation\;
\end{algorithm}
\bigskip

\section{Experimental analysis}
\label{sec:experimental}

\noindent We evaluate Pi-Transformer on five standard multivariate time-series anomaly detection benchmarks: SMD~\cite{SMD}, MSL~\cite{MSL}, SMAP~\cite{MSL}, SWaT~\cite{SWaT}, and PSM~\cite{PSM}. Table~\ref{tab:datasets} summarises the dataset domains and dimensionalities. We adopt this suite because it spans heterogeneous operating regimes where calibration varies and because it includes both magnitude-driven anomalies and timing/coordination disruptions, enabling evaluation of the complementary Energy and mismatch streams.\medskip

\begin{table}[!h]
\centering
\caption{Datasets used for benchmarking.}
\label{tab:datasets}
\setlength{\tabcolsep}{6pt}
\small
\begin{tabular}{l l r}
\toprule
Dataset & Domain & Channels ($C$) \\
\midrule
SMD  & Server / IT telemetry & 38 \\
MSL  & Spacecraft telemetry  & 55 \\
SMAP & Satellite telemetry   & 25 \\
SWaT & Industrial control (water treatment) & 51 \\
PSM  & Server / IT telemetry & 25 \\
\bottomrule
\end{tabular}
\end{table}

\noindent Competing methods include classical unsupervised detectors (LOF, OCSVM, IForest, VAR), probabilistic and subspace methods (MMPCACD, CL-MPPCA, BOCPD), deep generative and recurrent models (DAGMM, LSTM-VAE, BeatGAN, LSTM, OmniAnomaly, InterFusion, THOC), and recent sequence and attention-based detectors (U-Time, TS-CP2, Anomaly Transformer, DCDetector), together with ITAD. We report precision, recall, and F1-score for segment-level evaluation. All inputs are standardised per channel and split into overlapping windows of length $L$ with a stride of $1$. During inference, the model produces reconstruction error $r_i$, series-prior mismatch $\Delta_i$, prior-alignment weights $w_i$ from the time-wise softmax over $-\Delta$, and Energy $e_i = w_i r_i$. Robust normalisation with training set medians and interquartile ranges yields $\widetilde e_i$ and $\widetilde d_i$. The fused score $f_i = \max(\widetilde e_i, \widetilde d_i)$ is then thresholded at a single global percentile determined from fused scores on the training and threshold splits. 

\subsection{Benchmarking results}

Table~\ref{tab:benchmark} shows that Pi-Transformer achieves state-of-the-art or highly competitive F1 across the five benchmarks, with a balanced precision--recall profile. All comparisons in this section follow the segment-level point-adjust evaluation procedure described above and in Algorithm~\ref{alg:inference}. We note, however, that point-adjust is more permissive than stricter point-wise or range-aware metrics, because a detection within a labelled anomalous segment is credited across that segment. This distinction is relevant here because stride-1 overlapping windows and end-aligned score aggregation can produce slight lead or lag effects around anomaly boundaries. Accordingly, the results in Table~\ref{tab:benchmark} should be interpreted primarily as segment-detection performance under the adopted protocol, and absolute scores under stricter localisation-sensitive metrics may be lower.\medskip

\begin{table}[!h]
\centering
\caption{Benchmarking results (P/R/F1) across SMD, MSL, SMAP, SWaT, and PSM. Benchmark results are averaged over five seeds.}
\label{tab:benchmark}
\begin{adjustbox}{max width=\textwidth}
\begin{tabular}{l|ccc|ccc|ccc|ccc|ccc}
\toprule
\textbf{Dataset} & \multicolumn{3}{c|}{\textbf{SMD}} & \multicolumn{3}{c|}{\textbf{MSL}} & \multicolumn{3}{c|}{\textbf{SMAP}} & \multicolumn{3}{c|}{\textbf{SWaT}} & \multicolumn{3}{c}{\textbf{PSM}} \\
\hline
\textbf{Metric}  & P & R & F1 & P & R & F1 & P & R & F1 & P & R & F1 & P & R & F1 \\
\midrule
LOF & 56.34 & 39.86 & 46.68 & 47.72 & 85.25 & 61.18 & 58.93 & 56.33 & 57.60 & 72.15 & 65.43 & 68.62 & 57.89 & 90.49 & 70.61 \\
OCSVM & 44.34 & 76.72 & 56.19 & 59.78 & 86.87 & 70.82 & 53.85 & 59.07 & 56.34 & 45.39 & 49.22 & 47.23 & 62.75 & 80.89 & 70.67 \\
U-Time & 65.95 & 74.75 & 70.07 & 57.20 & 71.66 & 63.62 & 49.71 & 56.18 & 52.75 & 46.20 & 87.94 & 60.58 & 82.85 & 79.34 & 81.06 \\
IForest & 42.31 & 73.29 & 53.64 & 53.94 & 86.54 & 66.45 & 52.39 & 59.07 & 55.53 & 49.29 & 44.95 & 47.02 & 76.09 & 92.45 & 83.48 \\
DAGMM & 67.30 & 49.89 & 57.30 & 89.60 & 63.93 & 74.62 & 86.45 & 56.73 & 68.51 & 89.92 & 57.84 & 70.40 & 93.49 & 70.03 & 80.08 \\
ITAD & 86.22 & 73.71 & 79.48 & 69.44 & 84.09 & 76.07 & 82.42 & 66.89 & 73.85 & 63.13 & 52.08 & 57.08 & 72.80 & 64.02 & 68.13 \\
VAR & 78.35 & 70.26 & 74.08 & 74.68 & 81.42 & 77.90 & 81.38 & 53.88 & 64.83 & 81.59 & 60.29 & 69.34 & 90.71 & 83.82 & 87.13 \\
MMPCACD & 71.20 & 79.28 & 75.02 & 81.42 & 61.31 & 69.95 & 88.61 & 75.84 & 81.73 & 82.52 & 68.29 & 74.73 & 76.26 & 78.35 & 77.29 \\
CL-MPPCA & 82.36 & 76.07 & 79.09 & 73.71 & 88.54 & 80.44 & 86.13 & 63.16 & 72.88 & 76.78 & 81.50 & 79.07 & 56.02 & \textbf{99.93} & 71.80 \\
TS-CP2 & 87.42 & 66.25 & 75.38 & 86.45 & 68.48 & 76.42 & 87.65 & 83.18 & 85.36 & 81.23 & 74.10 & 77.50 & 82.67 & 78.16 & 80.35 \\
Deep-SVDD & 78.54 & 79.67 & 79.10 & 91.92 & 76.63 & 83.58 & 89.93 & 56.02 & 69.04 & 80.42 & 84.45 & 82.39 & 95.41 & 86.49 & 90.73 \\
BOCPD & 70.90 & 82.04 & 76.07 & 80.32 & 87.20 & 83.62 & 84.65 & 85.85 & 85.24 & 89.46 & 70.75 & 79.01 & 80.22 & 75.33 & 77.70 \\
LSTM-VAE & 75.76 & 90.08 & 82.30 & 85.49 & 79.94 & 82.62 & 92.20 & 67.75 & 78.10 & 76.00 & 89.50 & 82.20 & 73.62 & 89.92 & 80.96 \\
BeatGAN & 72.90 & 84.09 & 78.10 & 89.75 & 85.42 & 87.53 & 92.38 & 55.85 & 69.61 & 64.01 & 87.46 & 73.92 & 90.30 & 93.84 & 92.04 \\
LSTM & 78.55 & 85.28 & 81.78 & 85.45 & 82.50 & 83.95 & 89.41 & 78.13 & 83.39 & 86.15 & 83.27 & 84.69 & 76.93 & 89.64 & 82.80 \\
OmniAnomaly & 83.68 & 86.82 & 85.22 & 89.02 & 86.37 & 87.67 & 92.49 & 81.99 & 86.92 & 81.42 & 84.30 & 82.83 & 88.39 & 74.46 & 80.83 \\
InterFusion & 87.02 & 85.43 & 86.22 & 81.28 & 92.70 & 86.62 & 89.77 & 88.52 & 89.14 & 80.59 & 85.58 & 83.01 & 83.61 & 83.45 & 83.52 \\
THOC & 79.76 & 90.95 & 84.99 & 88.45 & 90.97 & 89.69 & 92.06 & 89.34 & 90.68 & 83.94 & 86.36 & 85.13 & 88.14 & 90.99 & 89.54 \\
AnomalyTrans & \textbf{88.47} & 92.28 & 90.33 & 91.92 & 96.03 & 93.93 & 93.59 & \textbf{99.41} & 96.41 & 89.10 & 99.28 & 94.22 & 96.94 & 97.81 & 97.37 \\
DCDetector & 83.59 & 91.10 & 87.18 & 93.69 & \textbf{99.69} & \textbf{96.60} & 95.63 & 98.92 & \textbf{97.02} & 93.11 & 99.77 & 96.33 & 97.14 & 98.74 & 97.94 \\
Pi-Transformer (ours) & 88.19 & \textbf{94.49} & \textbf{91.23} & \textbf{96.24} & 95.69 & 95.96 & \textbf{96.81} & 97.24 & \textbf{97.02} & \textbf{93.84} & \textbf{100.00} & \textbf{96.82} & \textbf{97.37} & 98.80 & \textbf{98.08} \\
\bottomrule
\end{tabular}
\end{adjustbox}
\end{table}

\begin{figure}[!h]
\centering
\includegraphics[width=\textwidth]{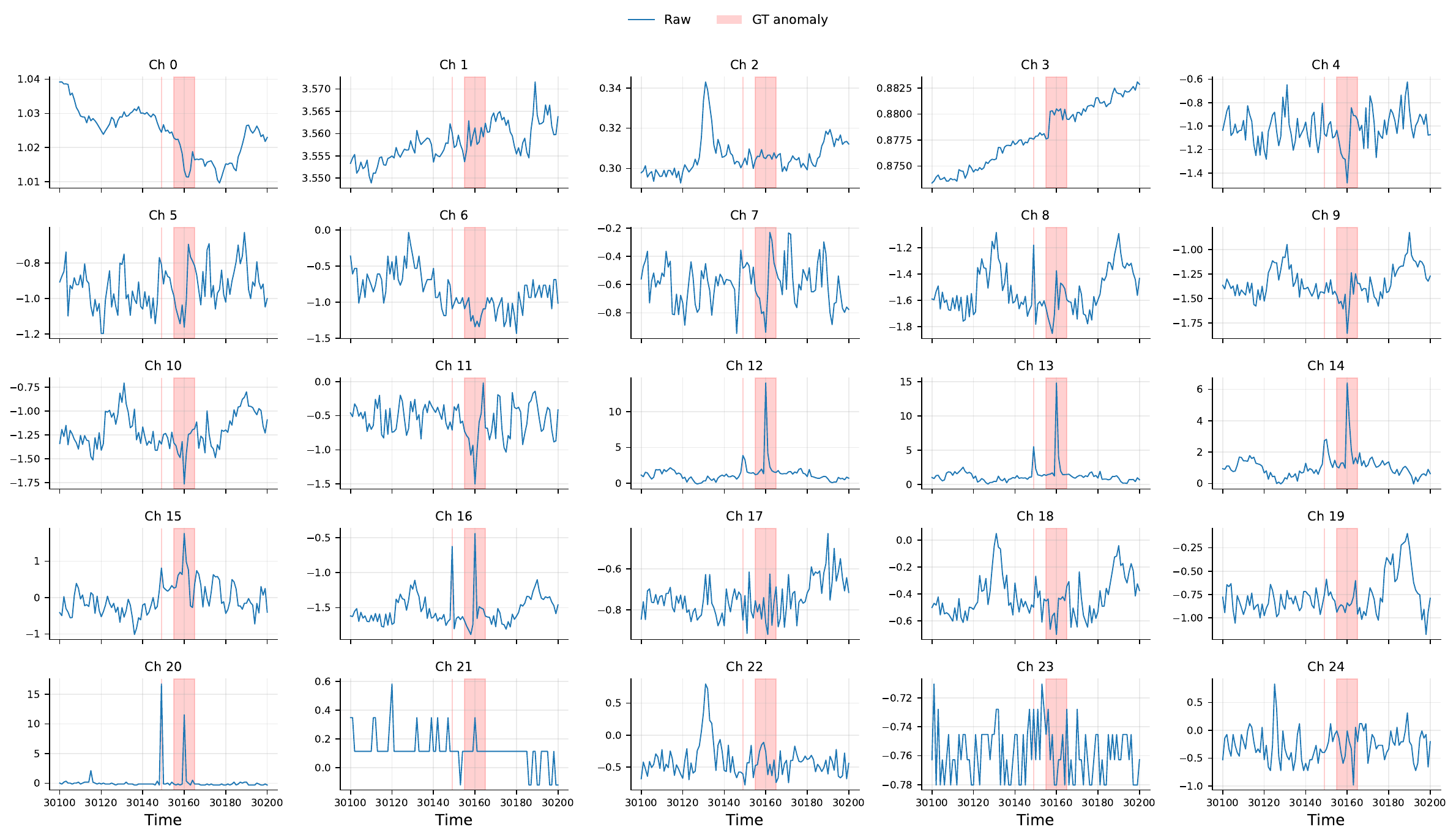}	
\caption{\textbf{PSM example (input window).} Twenty-five observed channels within a representative window from the PSM dataset. Two ground-truth anomalous segments are shaded. Several channels show clear deviations aligned with the later regime, while others remain near baseline.}
\label{fig:psm_window}
\end{figure}

\noindent To illustrate these mechanisms, we analyse outputs on the PSM dataset. Figure~\ref{fig:psm_window} presents an input window consisting of 25 observed channels. Two ground-truth anomalous regimes are present within this window (shaded). Several channels (e.g., Ch~13/14/20) show clear amplitude deviations aligned with the later regime, while others remain near baseline, underscoring the multivariate and partially correlated nature of the task.\medskip

\begin{figure}[!h]
\centering
\includegraphics[width=\textwidth]{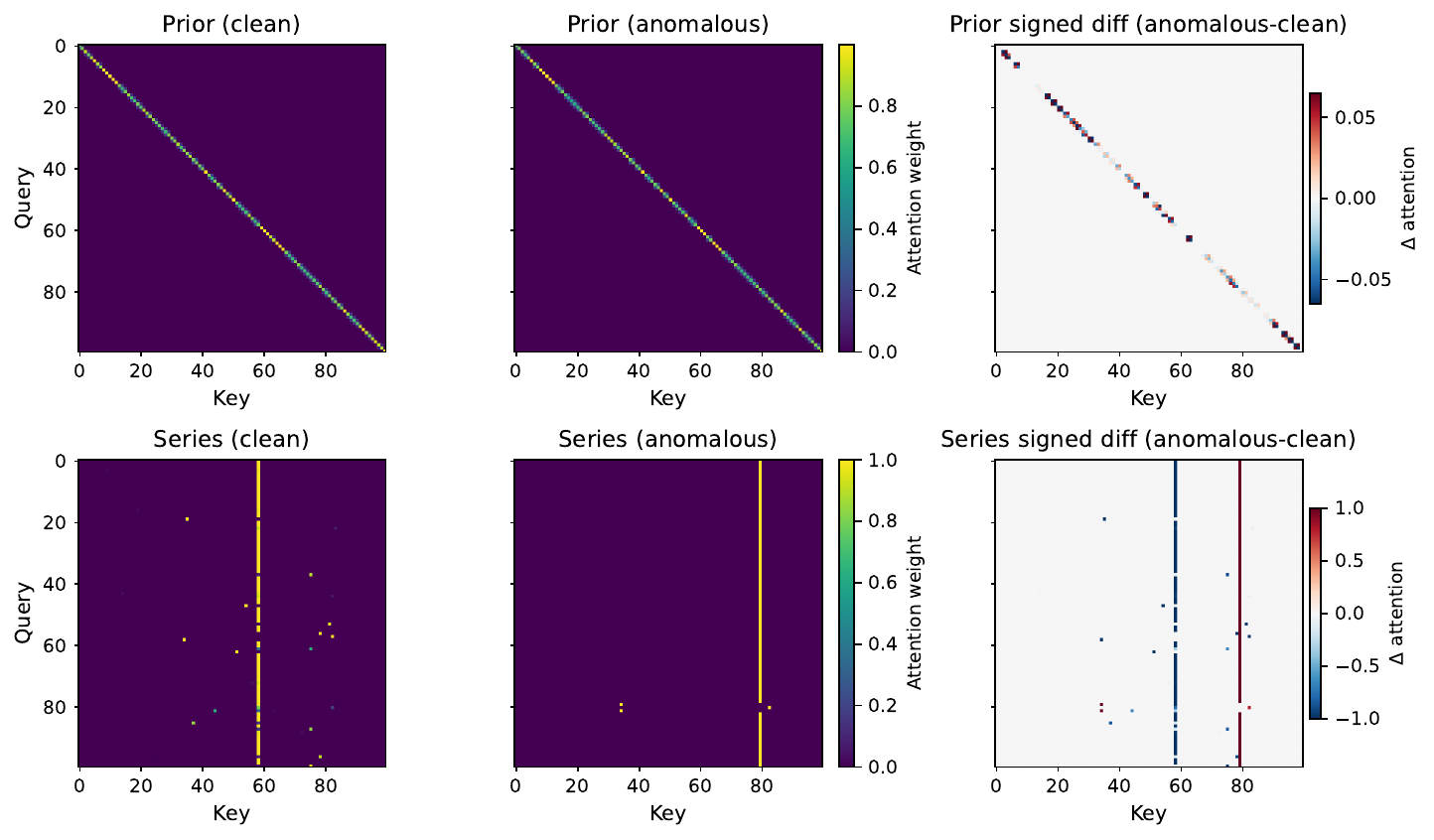}
\caption{\textbf{PSM example (attention maps).} Prior and series attention under a clean window (left) and an anomalous window (middle), with the signed difference (anomaly$-$clean; right). Top row: prior attention. Bottom row: series attention. The anomalous regime induces a clearer reallocation in the series pathway, while the prior remains comparatively stable.}
\label{fig:psm_attn}
\end{figure}

\noindent Figure~\ref{fig:psm_attn} visualises the corresponding attention behaviour using a representative clean window and an anomalous window. The left and middle columns show the prior and series attention maps under clean and anomalous conditions, while the right column shows the signed difference (anomaly$-$clean). The series pathway exhibits a pronounced reallocation under the anomalous regime, whereas the prior remains smoother and changes more conservatively. This qualitative separation is consistent with the mechanism summarised in Figure~\ref{fig:pitransattention} in the sense that phase-breaking events are characterised by increased divergence between series and prior attention, which is reflected in an elevated mismatch response.\medskip

\begin{figure}[!h]
\centering
\includegraphics[width=\textwidth]{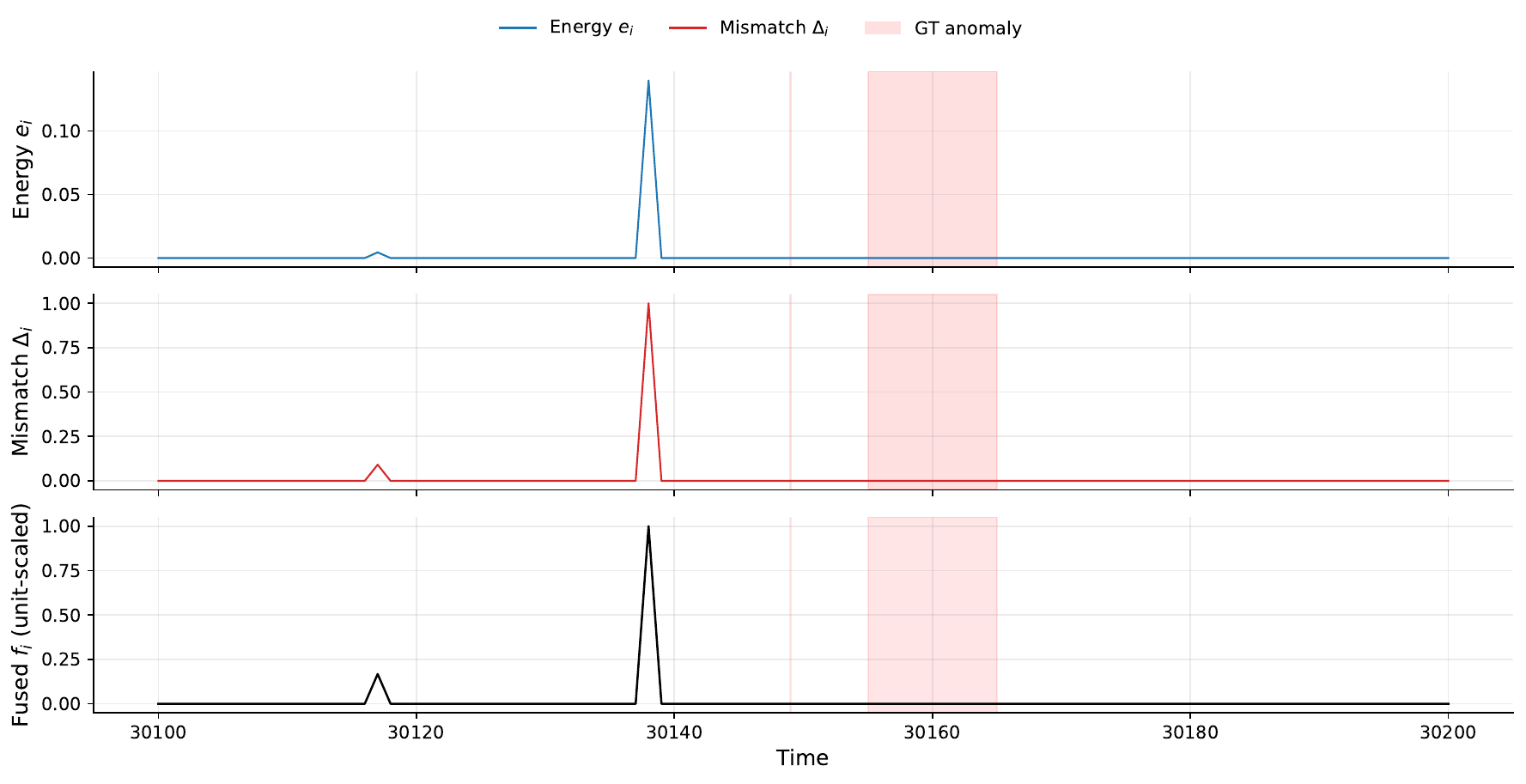}	
\caption{\textbf{PSM example (decision streams).}
From top to bottom: Energy $e_i$, mismatch $\Delta_i$ (raw, end-anchored), and fused score $f_i=\max(\widetilde e_i,\widetilde d_i)$ (unit-scaled) for the same window as Fig.~\ref{fig:psm_window}. 
Two ground-truth anomaly segments are shaded. Energy peaks just before the second anomalous onset, while being muted exactly at the breakpoint where prior-alignment weights collapse.}

\label{fig:psm_window_signals}
\end{figure}

\noindent For the same window as Figure~\ref{fig:psm_window}, Figure~\ref{fig:psm_window_signals} shows the decision streams. Around the earlier, shorter regime, both $e_i$ and $\Delta_i$ exhibit a modest rise near its leading edge. Before the later, longer regime, a pronounced pre-onset spike appears in both $e_i$ and $\Delta_i$; with stride-1, end-anchored windows this produces a lead effect, i.e., peaks just before the shaded span. At the breakpoint of the longer regime, prior-alignment weights can collapse, which can mute $e_i$ inside the shaded region. $\Delta_i$ typically spikes at the disruption and then decays as the window becomes dominated by the new regime. The fused score $f_i$ therefore fires strongly at the leading edges of both regimes and need not remain elevated throughout the interior.\medskip

\begin{figure}[!h]
\centering
\includegraphics[width=\textwidth]{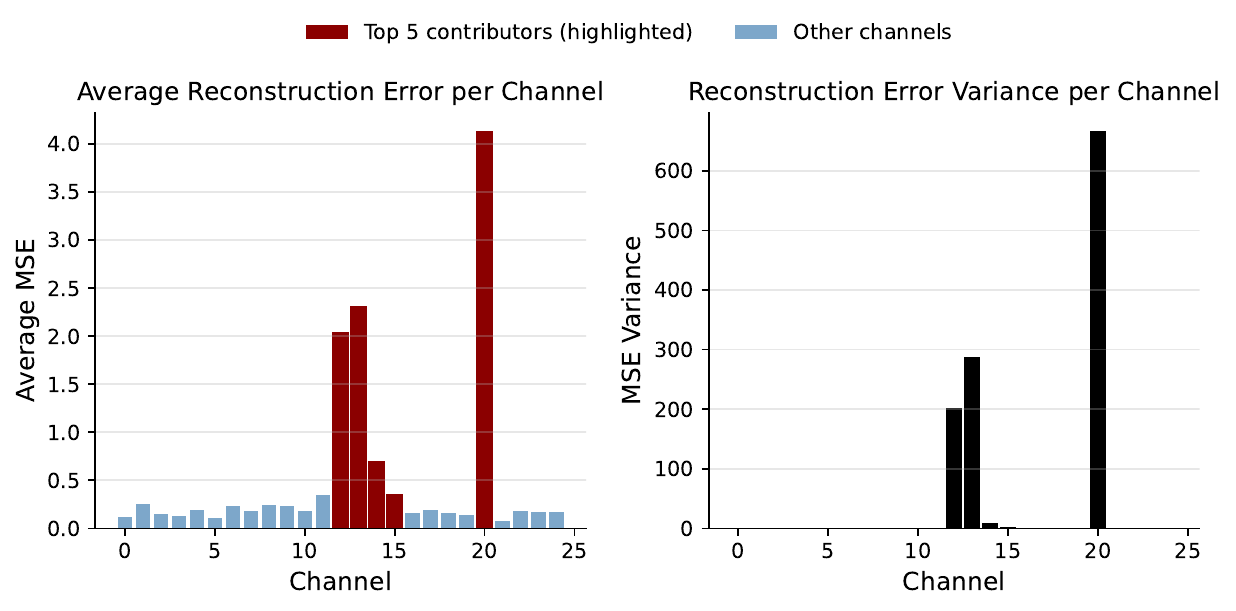}	
\caption{\textbf{PSM example (channel contributions).} Average (left) and variance (right) of reconstruction error per channel within the same window as Fig.~\ref{fig:psm_window}. Errors concentrate in a subset of channels whose dynamics lose synchrony during the anomalous regime, with the top contributors highlighted.}
\label{fig:psm_window_analysis}
\end{figure}

\noindent Figure~\ref{fig:psm_window_analysis} shows channel contributions within this window. The left panel shows the average reconstruction error per channel, while the right panel displays the error variance per channel. Although individual channels may appear normal in isolation, their joint phase relationships break down around the annotated regimes, leading to systematic elevation in reconstruction error. This highlights that anomaly evidence is not confined to visibly deviating channels, but is distributed according to both magnitude and coordination across the multivariate series.\medskip

\bigskip

\subsection{Hyperparameter sensitivity and ablations}

\noindent We analyse the sensitivity of Pi-Transformer to architectural and optimisation choices and quantify the contribution of its main components. Unless otherwise stated, each ablation changes one factor while keeping the remaining settings identical to the full model.\medskip

\begin{table}[!h]
\centering
\caption{\textbf{Ablation results across datasets.} Metrics are accuracy, precision, recall, and F1. The table reports sensitivity to model and training hyperparameters and component ablations of Pi-Transformer. ``Prior pathway: Full'' denotes the complete model, ``No prior'' removes the prior pathway, and ``Single-head prior'' restricts the prior to one head. ``Prior Components'' removes individual elements of the learned prior, ``Prior Replacement'' substitutes the learned prior with fixed alternatives, and ``Score Stream'' evaluates Energy-only, Mismatch-only, and the fused score.}
\label{tab:ablation}
\begin{adjustbox}{max width=\textwidth}
\begin{tabular}{l|c|cccc|cccc|cccc|cccc|cccc}
\toprule
\multirow{2}{*}{Parameter} & \multirow{2}{*}{Value} & \multicolumn{4}{c|}{MSL} & \multicolumn{4}{c|}{PSM} & \multicolumn{4}{c|}{SMAP} & \multicolumn{4}{c|}{SMD} & \multicolumn{4}{c}{SWaT} \\
\cmidrule(lr){3-6} \cmidrule(lr){7-10} \cmidrule(lr){11-14} \cmidrule(lr){15-18} \cmidrule(lr){19-22}
 & & Acc & Prec & Rec & F1 & Acc & Prec & Rec & F1 & Acc & Prec & Rec & F1 & Acc & Prec & Rec & F1 & Acc & Prec & Rec & F1 \\
\midrule
\multirow{3}{*}{Prior Pathway} & Full & 99.15 & 96.24 & 95.69 & 95.96 & 98.93 & 97.37 & 98.80 & 98.08 & 99.24 & 96.81 & 97.24 & 97.02 & 99.25 & 88.19 & 94.49 & 91.23 & 99.20 & 93.84 & 100.00 & 96.82 \\
 & No Prior & 94.67 & 84.37 & 60.70 & 70.61 & 88.93 & 73.24 & 94.68 & 82.59 & 93.90 & 95.97 & 54.58 & 69.59 & 97.73 & 67.14 & 88.96 & 76.52 & 58.11 & 21.06 & 89.20 & 34.08 \\
 & Single-head Prior& 98.21 & 95.36 & 87.30 & 91.15 & 98.70 & 97.31 & 98.03 & 97.67 & 99.16 & 96.22 & 97.21 & 96.71 & 99.24 & 87.66 & 94.97 & 91.17 & 81.92 & 39.75 & 94.93 & 56.04 \\
\midrule
\multirow{3}{*}{Enc. Layers} & 2 & 98.27 & 95.30 & 87.96 & 91.48 & 98.93 & 97.37 & 98.80 & 98.08 & 99.24 & 96.81 & 97.24 & 97.02 & 99.25 & 88.19 & 94.49 & 91.23 & 98.25 & 90.08 & 96.22 & 93.05 \\
 & 3 & 99.15 & 96.24 & 95.69 & 95.96 & 98.82 & 97.41 & 98.36 & 97.88 & 98.30 & 96.66 & 89.85 & 93.13 & 99.04 & 87.13 & 90.23 & 88.65 & 99.20 & 93.84 & 100.00 & 96.82 \\
 & 4 & 97.60 & 95.47 & 81.07 & 87.68 & 98.84 & 97.42 & 98.41 & 97.91 & 98.15 & 96.69 & 88.58 & 92.46 & 98.86 & 86.12 & 86.55 & 86.34 & 99.20 & 93.85 & 100.00 & 96.83 \\
\midrule
\multirow{4}{*}{Model Dim.} & 128 & 98.63 & 95.62 & 91.18 & 93.35 & 98.52 & 97.31 & 97.35 & 97.33 & 94.52 & 95.80 & 59.78 & 73.62 & 98.97 & 85.98 & 89.75 & 87.82 & 99.12 & 93.28 & 100.00 & 96.52 \\
 & 256 & 98.43 & 95.56 & 89.24 & 92.29 & 98.80 & 97.37 & 98.31 & 97.84 & 94.79 & 95.94 & 61.89 & 75.24 & 99.05 & 86.88 & 90.87 & 88.83 & 99.04 & 93.77 & 98.68 & 96.16 \\
 & 512 & 99.15 & 96.24 & 95.69 & 95.96 & 98.93 & 97.37 & 98.80 & 98.08 & 99.24 & 96.81 & 97.24 & 97.02 & 99.25 & 88.19 & 94.49 & 91.23 & 99.20 & 93.84 & 100.00 & 96.82 \\
 & 1024 & 99.08 & 95.47 & 95.85 & 95.66 & 98.89 & 97.43 & 98.59 & 98.00 & 95.53 & 95.65 & 68.14 & 79.58 & 99.14 & 87.44 & 92.71 & 90.00 & 99.21 & 93.89 & 100.00 & 96.85 \\
\midrule
\multirow{4}{*}{Num. Heads} & 2 & 98.25 & 95.48 & 87.54 & 91.33 & 98.80 & 97.06 & 98.67 & 97.86 & 97.14 & 96.29 & 80.78 & 87.85 & 99.15 & 88.33 & 91.77 & 90.02 & 99.20 & 93.84 & 100.00 & 96.82 \\
 & 4 & 98.48 & 95.61 & 89.72 & 92.57 & 98.74 & 97.19 & 98.31 & 97.75 & 94.32 & 95.80 & 58.18 & 72.40 & 99.16 & 87.95 & 92.53 & 90.18 & 99.20 & 93.78 & 100.00 & 96.79 \\
 & 8 & 99.15 & 96.24 & 95.69 & 95.96 & 98.93 & 97.37 & 98.80 & 98.08 & 99.24 & 96.81 & 97.24 & 97.02 & 99.25 & 88.19 & 94.49 & 91.23 & 99.20 & 93.84 & 100.00 & 96.82 \\
 & 16 & 98.27 & 95.60 & 87.63 & 91.44 & 98.93 & 97.29 & 98.88 & 98.08 & 99.00 & 96.41 & 95.76 & 96.08 & 99.34 & 88.33 & 96.90 & 92.42 & 99.12 & 93.25 & 100.00 & 96.51 \\
\midrule
\multirow{4}{*}{Batch Size} & 64 & 98.21 & 95.93 & 86.71 & 91.09 & 98.87 & 97.42 & 98.54 & 97.98 & 96.49 & 96.56 & 75.26 & 84.59 & 99.34 & 88.39 & 96.95 & 92.48 & 99.22 & 93.95 & 100.00 & 96.88 \\
 & 128 & 97.77 & 94.83 & 83.39 & 88.74 & 98.93 & 97.46 & 98.72 & 98.08 & 98.94 & 96.44 & 95.22 & 95.82 & 99.27 & 88.45 & 94.78 & 91.51 & 99.18 & 93.70 & 100.00 & 96.75 \\
 & 256 & 99.15 & 96.24 & 95.69 & 95.96 & 98.93 & 97.37 & 98.80 & 98.08 & 93.82 & 95.30 & 54.37 & 69.24 & 99.25 & 88.19 & 94.49 & 91.23 & 99.20 & 93.84 & 100.00 & 96.82 \\
 & 512 & 97.68 & 95.13 & 82.17 & 88.17 & 98.85 & 97.42 & 98.45 & 97.93 & 99.24 & 96.81 & 97.24 & 97.02 & 99.36 & 88.95 & 96.63 & 92.63 & 99.13 & 93.30 & 100.00 & 96.53 \\
\midrule
\multirow{3}{*}{Epochs} & 3 & 97.72 & 94.95 & 82.73 & 88.42 & 98.93 & 97.37 & 98.80 & 98.08 & 95.65 & 95.88 & 68.98 & 80.23 & 99.28 & 87.83 & 95.87 & 91.67 & 99.20 & 93.84 & 100.00 & 96.82 \\
 & 5 & 99.15 & 96.24 & 95.69 & 95.96 & 98.87 & 97.11 & 98.85 & 97.97 & 99.24 & 96.81 & 97.24 & 97.02 & 99.18 & 88.02 & 92.87 & 90.38 & 99.16 & 93.51 & 100.00 & 96.65 \\
 & 10 & 98.17 & 95.06 & 87.15 & 90.93 & 98.82 & 97.23 & 98.55 & 97.88 & 97.52 & 96.25 & 83.85 & 89.62 & 99.25 & 88.19 & 94.49 & 91.23 & 99.21 & 93.87 & 100.00 & 96.84 \\
 \midrule
 \multirow{6}{*}{Prior Components}
 & w/o phase gate
& 98.26 & 95.50 & 87.66 & 91.41
& 98.90 & 97.24 & 98.83 & 98.03
& 98.57 & 97.30 & 90.70 & 94.19
& 99.32 & 88.86 & 95.78 & 92.20
& 99.21 & 93.86 & 100.00 & 96.83 \\
 & w/o fractal warping
& 97.69 & 95.31 & 82.09 & 88.20
& 98.87 & 97.15 & 98.81 & 97.97
& 95.98 & 96.13 & 72.34 & 81.92
& 99.32 & 88.82 & 95.91 & 92.23
& 99.19 & 93.77 & 100.00 & 96.78 \\
 & w/o Gaussian time kernel
& 97.18 & 95.37 & 76.93 & 85.16
& 98.82 & 97.41 & 98.36 & 97.88
& 94.87 & 95.99 & 65.74 & 76.57
& 99.00 & 87.22 & 88.90 & 88.05
& 99.14 & 93.41 & 100.00 & 96.59 \\
 & w/o Hurst distillation
& 98.05 & 95.55 & 85.45 & 90.22
& 98.92 & 97.25 & 98.90 & 98.07
& 95.94 & 96.16 & 71.90 & 81.67
& 99.29 & 88.67 & 94.99 & 91.72
& 99.20 & 93.79 & 100.00 & 96.79 \\
 & w/o smoothness
& 98.53 & 95.79 & 89.99 & 92.80
& 98.84 & 97.16 & 98.69 & 97.92
& 95.94 & 96.16 & 71.94 & 81.71
& 99.34 & 88.92 & 96.04 & 92.34
& 99.21 & 93.87 & 100.00 & 96.84 \\
\midrule
\multirow{3}{*}{Prior Replacement} & Uniform
& 97.26 & 95.22 & 76.81 & 85.03
& 98.83 & 97.12 & 98.73 & 97.92
& 98.27 & 97.15 & 97.86 & 97.50
& 98.35 & 84.04 & 90.65 & 87.22
& 98.44 & 89.51 & 98.68 & 93.87 \\
 & Exponential
& 97.50 & 94.70 & 80.75 & 87.17
& 98.83 & 97.12 & 98.73 & 97.92
& 98.71 & 98.19 & 98.42 & 98.30
& 98.42 & 84.68 & 90.65 & 87.56
& 99.19 & 93.71 & 100.00 & 96.75 \\
 & Fixed Gaussian
& 98.51 & 95.57 & 89.99 & 92.70
& 98.79 & 97.11 & 98.51 & 97.80
& 98.39 & 97.16 & 98.06 & 97.61
& 99.32 & 88.82 & 95.91 & 92.23
& 99.20 & 93.83 & 100.00 & 96.82 \\
\midrule
\multirow{3}{*}{Score Stream} & Energy only
& 97.94 & 95.09 & 84.82 & 89.66
& 98.83 & 97.12 & 98.73 & 97.92
& 95.97 & 96.17 & 71.34 & 81.91
& 99.32 & 88.77 & 95.84 & 92.17
& 99.15 & 93.44 & 100.00 & 96.61 \\
 & Mismatch only
& 97.08 & 94.74 & 76.58 & 84.70
& 97.54 & 97.44 & 93.57 & 95.47
& 93.68 & 95.67 & 52.98 & 68.19
& 98.16 & 85.20 & 67.55 & 75.36
& 89.01 & 86.23 & 11.28 & 19.95 \\
 & Fused
& 99.15 & 96.24 & 95.69 & 95.96 
& 98.93 & 97.37 & 98.80 & 98.08 
& 99.24 & 96.81 & 97.24 & 97.02 
& 99.25 & 88.19 & 94.49 & 91.23 
& 99.20 & 93.84 & 100.00 & 96.82 \\
\bottomrule
\end{tabular}
\end{adjustbox}
\end{table}

\noindent The results show that the prior pathway is the dominant contributor to performance across all benchmarks. Table~\ref{tab:ablation} (Prior Pathway) shows removing the prior stream produces the largest degradation in F1 on every dataset, indicating that reconstruction evidence alone is not reliably calibrated under regime changes and cross-variable coordination. Restricting the prior to a single head retains partial benefit but remains consistently below the full multi-head prior, supporting the role of multiple heads in representing a richer set of nominal timing templates and improving robustness to benign temporal variability.\medskip

\noindent The architectural sweeps indicate that strong performance is achieved in a moderate-capacity regime. Varying encoder depth, model dimension, and number of attention heads yields diminishing returns beyond mid-range settings, and in some cases larger configurations slightly reduce F1. This behaviour is consistent with the detector relying on stable attention structure and calibrated scoring rather than purely increased representational capacity. The number of heads is also tied to the prior pathway expressiveness: moderate multi-head designs are effective, whereas overly small head counts reduce stability and overly large head counts yield limited additional gain.\medskip

\noindent Optimisation ablations over batch size and training epochs indicate that the method is not overly sensitive to the training schedule within the tested range. Moderate batch sizes and short-to-moderate training budgets already achieve strong results, while larger batches or longer training provide limited improvement and can shift the precision--recall balance. This is consistent with the role of the prior pathway as a structural reference: once a stable alignment between series and prior attention is established, further reduction of reconstruction loss yields decreasing marginal benefit for detection.\medskip

\noindent We further isolates the internal design choices behind the prior attention and the test-time scoring. In the Prior Components block, the Gaussian temporal kernel provides the most consistent benefit. Removing it yields the largest and most systematic F1 reductions across datasets, indicating that enforcing temporally localised influence is central to constructing a useful reference prior. Removing the phase gate, fractal warping, or the smoothness and distillation regularisers produces smaller but non-negligible changes, and their impact is dataset-dependent, which is expected because the prevalence of timing and coordination anomalies differs across benchmarks. The Prior Replacement block confirms that fixed priors can be competitive and even outperform the learned prior on some datasets, but performance is not consistent across benchmarks and the learned prior is the most reliable overall. These results support learning a structured prior and regularising it to evolve smoothly over time rather than relying on a generic decay alone.\medskip

\noindent The Score Stream block separates the two inference signals. The first is the alignment-weighted reconstruction score (Energy) and the second is the series--prior mismatch score (Mismatch). Energy-only is generally stronger because many anomalies appear as amplitude or shape deviations that reconstruction error captures directly. Mismatch-only can be substantially weaker on some datasets (for example, SWaT), which is consistent with mismatch behaving as a sparse indicator that activates mainly when anomalies disrupt timing or coordination rather than magnitude. Even when mismatch is weak as a standalone detector, it can still provide complementary evidence and improve the fused score on datasets where timing disruptions are prominent (for example, MSL and SMAP). Under fusion, mismatch can raise anomaly points that would otherwise remain below the decision threshold under Energy-only.\medskip

\noindent On SMD, the fused score is slightly lower than Energy-only. This behaviour is expected under the adopted normalisation and percentile thresholding applied to the fused score distribution. Although fusion never reduces scores pointwise, the addition of mismatch-induced high scores changes the overall score distribution and can increase the percentile threshold. On SMD this shifts the operating point towards higher precision but lower recall, yielding a small net F1 decrease. This suggests that, for SMD, the dominant anomaly morphology is already well captured by reconstruction evidence and mismatch contributes less aligned signal under the benchmark annotations.\medskip

\begin{figure}[!h]
    \centering
    \includegraphics[width=\textwidth]{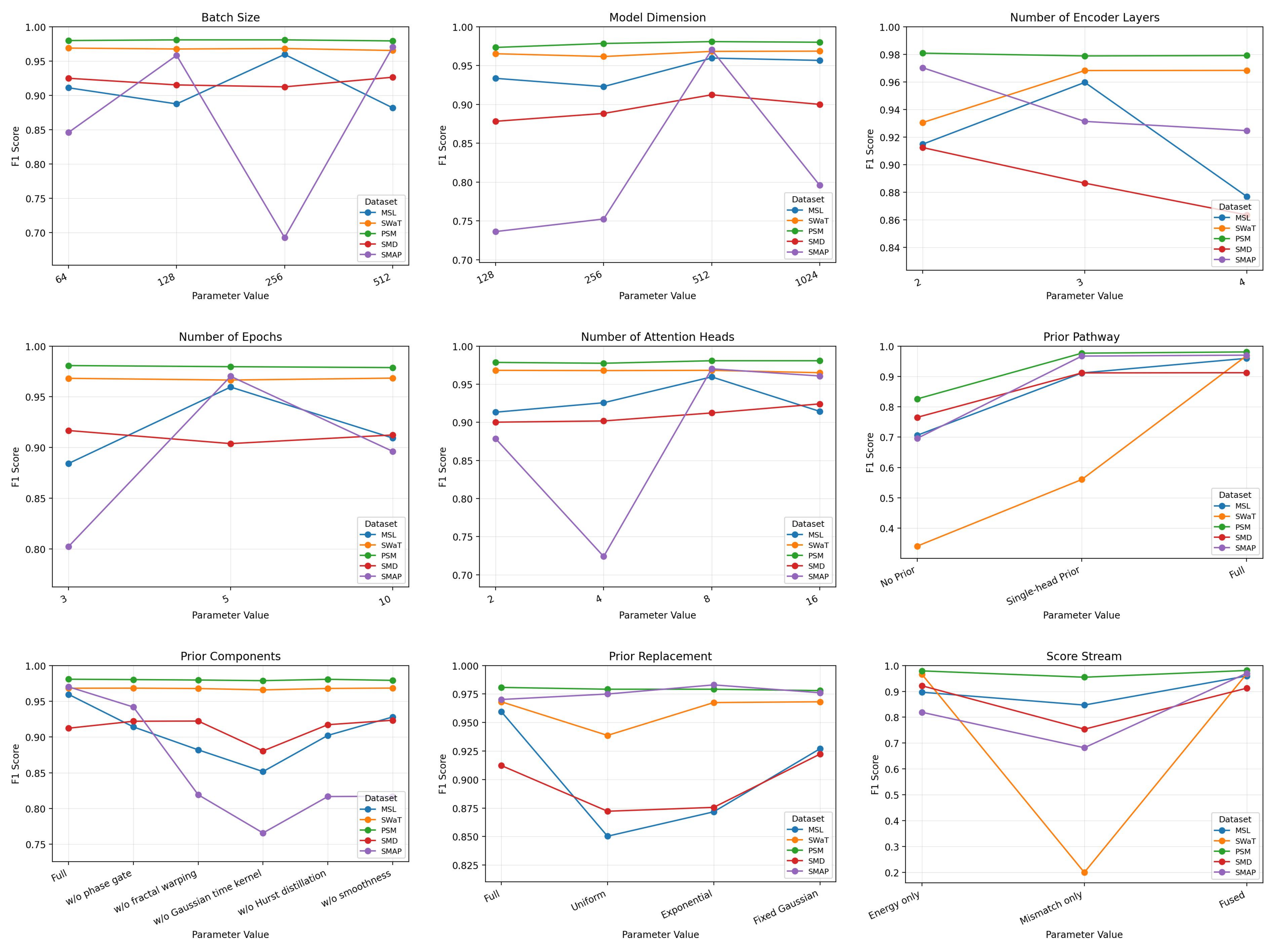}
    \caption{\textbf{Ablation summary.} F1 scores across five datasets (MSL, PSM, SMAP, SMD, SWaT) under variations of batch size, model dimension, encoder depth, training epochs, number of attention heads, and phase synchrony. Each panel reports performance for one factor, with coloured curves denoting datasets.}
    \label{fig:ablation}
\end{figure}

\noindent Figure~\ref{fig:ablation} summarises the hyperparameter sweeps and reinforces these conclusions. Across datasets, the prior pathway provides the largest and most consistent gains, multi-head priors are more stable than single-head variants, and moderate depth and width are sufficient to reach peak performance. Dataset-specific trends align with differences in anomaly characteristics: SWaT is most sensitive to removing the prior pathway, MSL and SMAP benefit strongly from incorporating timing and coordination evidence, and SMD is more dominated by reconstruction-driven detection under the adopted fusion and thresholding scheme. Lastly, localisation hyperparameters behave as expected. Increasing the inverse-temperature $\mathcal{T}$ in Eq.~\eqref{eq:Delta} sharpens the prior-alignment softmax of Eq.~\eqref{eq:softmax_w}, and window length $L$ trades context for localisation, with stride~1 overlap introducing a predictable lead effect that can be mitigated by right-edge attribution or aggregation across covering windows.

\noindent Additionally we study the sensitivity to the divergence inverse-temperature $\mathcal{T}$ via a small log-scale sweep centred at the dataset-specific optimum $\mathcal{T}^\star$. Concretely, we evaluate

\[
\mathcal{T} \in \{\mathcal{T}^\star/4,\;\mathcal{T}^\star/2,\;\mathcal{T}^\star,\;2\mathcal{T}^\star,\;4\mathcal{T}^\star\}\;.
\]

\noindent We report the corresponding endpoints $\mathcal{T}_{\min}=\mathcal{T}^\star/4$ and $\mathcal{T}_{\max}=4\mathcal{T}^\star$.\medskip

\noindent Table~\ref{tab:temp_sweep2} reports $\mathcal{T}_{\min}$, $\mathcal{T}_{\max}$, and $\mathcal{T}^\star$ together with the best F1 and the sweep spread $\Delta$F1.
Across datasets, the variation is small relative to the absolute F1, with the largest spread occurring on SMAP ($\Delta$F1$=0.325$) and near-flat behaviour on PSM ($\Delta$F1$=0.000$).\medskip

\begin{table}[!h]
\centering
\caption{\textbf{Sensitivity to divergence inverse-temperature $\mathcal{T}$.} $\mathcal{T}^\star$ is the best setting in the sweep. $\Delta\mathrm{F1}$ is the difference between the best and worst F1 across the sweep.}
\label{tab:temp_sweep2}
\begin{tabular}{lrrrrr}
\toprule
Dataset & $\mathcal{T}_{\min}$ & $\mathcal{T}_{\max}$ & $\mathcal{T}^\star$ & F1 at $\mathcal{T}^\star$ & $\Delta$F1 \\
\midrule
MSL  & 49.00 & 784.00 & 196.00 & 95.96 & 0.087 \\
PSM  & 23.94 & 382.96 & 95.74 & 98.08 & 0.000 \\
SMAP & 11.69 & 186.97 & 46.74 & 97.02 & 0.325 \\
SMD  & 42.58 & 681.35 & 170.33 & 91.23 & 0.005 \\
SWaT & 34.34 & 549.37 & 137.34 & 96.82 & 0.001 \\
\bottomrule
\end{tabular}
\end{table}

\noindent We also study the sensitivity to two prior-specific scalars, namely the coupling strength $k$ that scales the attention divergence term and the smoothness regulariser weight $\lambda_{\mathrm{smooth}}$ in the prior pathway. For each dataset we apply a small multiplicative sweep around the dataset default value. For $k$ we evaluate $\{0.5k_0,\;k_0,\;2k_0 ,\; 4k_0 \}$, where $k_0$ is the default setting. For $\lambda_{\mathrm{smooth}}$ we evaluate $\{0.5\lambda_0,\;\lambda_0,\;2\lambda_0\}$, where $\lambda_0$ is the default setting.\medskip

\noindent Table~\ref{tab:scalar_sweep} reports the tested range, the best setting, and the sweep spread $\Delta$F1.
Across datasets, the variation is small for PSM and SWaT, while SMAP shows the largest spread for both parameters.\medskip

\begin{table}[!h]
\centering
\caption{\textbf{Sensitivity to the coupling strength $k$ and smoothness weight $\lambda_{\mathrm{smooth}}$.} For each dataset, $\Delta\mathrm{F1}$ is the difference between the best and worst F1 across the sweep.}
\label{tab:scalar_sweep}
\begin{tabular}{lrrrrr}
\toprule
Dataset & Min & Max & Best & F1 at Best & $\Delta$F1 \\
\midrule
\multicolumn{6}{l}{Coupling strength $k$} \\
\midrule
MSL  & 2  & 16 & 8  & 95.96 & 0.049 \\
PSM  & 2  & 20 & 2  & 98.08 & 0.005 \\
SMAP & 1  & 8  & 8  & 97.02 & 0.159 \\
SMD  & 1  & 4  & 1  & 91.23 & 0.018 \\
SWaT & 1  & 4  & 2  & 96.82 & 0.003 \\
\midrule
\multicolumn{6}{l}{Smoothness weight $\lambda_{\mathrm{smooth}}$} \\
\midrule
MSL  & 0.000070 & 0.000279 & 0.000279 & 95.96 & 0.051 \\
PSM  & 0.000057 & 0.000229 & 0.000229 & 98.08 & 0.005 \\
SMAP & 0.000266 & 0.001063 & 0.000266 & 97.02 & 0.144 \\
SMD  & 0.000094 & 0.000377 & 0.000377 & 91.23 & 0.031 \\
SWaT & 0.020070 & 0.080279 & 0.080279 & 96.82 & 0.057 \\
\bottomrule
\end{tabular}
\end{table}

\noindent We evaluate computational cost on a single NVIDIA Tesla T4. Table~\ref{tab:efficiency_dmodel_all} reports training-time efficiency as the averaged per-iteration running time measured over 100 training iterations (Time) and the corresponding peak GPU memory footprint during training (Mem), for different model widths. Results are reported alongside Anomaly Transformer~\cite{xu2021anomaly} and DCDetector~\cite{yang2023dcdetector}. Table~\ref{tab:temp_runtime} additionally reports end-to-end wall-clock time for training and evaluation to reflect full pipeline cost.\medskip

\noindent In terms of scaling, for window length $L$, model width $d_{\text{model}}$, $N$ encoder layers, and $H$ attention heads with head dimension $d_h=d_{\text{model}}/H$, the dominant cost per forward pass is standard multi-head self-attention, $\mathcal{O}(NH L^2 d_h)=\mathcal{O}(N L^2 d_{\text{model}})$. The prior pathway constructs structured kernels over the same $L\times L$ grid and adds $\mathcal{O}(NH L^2)$ operations; thus, the asymptotic complexity matches a conventional Transformer up to constant factors.

\begin{table}[!h]
\centering
\caption{End-to-end wall-clock runtime on a single NVIDIA Tesla T4.}
\label{tab:temp_runtime}
\setlength{\tabcolsep}{6pt}
\small
\begin{tabular}{lrr}
\toprule
Dataset & Train time (s) & Eval time (s) \\
\midrule
MSL  & 29.15  & 0.45 \\
PSM  & 34.09  & 0.39 \\
SMAP & 34.20  & 1.44 \\
SMD  & 176.76 & 2.97 \\
SWaT & 250.58 & 3.21 \\
\bottomrule
\end{tabular}
\end{table}
\medskip

\begin{table}[!h]
\centering
\caption{Training-time efficiency comparison on a single NVIDIA Tesla T4: averaged running time per iteration over 100 training iterations (Time) and peak GPU memory footprint during training (Mem), reported under different $d_{\text{model}}$ for Anomaly Transformer~\cite{xu2021anomaly}, DCDetector~\cite{yang2023dcdetector}, and Pi-Transformer.}
\label{tab:efficiency_dmodel_all}
\setlength{\tabcolsep}{3pt}
\small
\begin{tabular}{c cc cc cc}
\toprule
& \multicolumn{2}{c}{Anomaly Transformer} & \multicolumn{2}{c}{DCDetector} & \multicolumn{2}{c}{Pi-Transformer} \\
\cmidrule(lr){2-3}\cmidrule(lr){4-5}\cmidrule(lr){6-7}
$d_{\text{model}}$ & Mem (GB) & Time (s/iter) & Mem (GB) & Time (s/iter) & Mem (GB) & Time (s/iter) \\
\midrule
128  & --  & --   & 3.9  & 0.05 & 4.8  & 0.12 \\
256  & 4.9 & 0.12 & 6.1  & 0.10 & 5.5  & 0.20 \\
512  & 5.5 & 0.15 & 10.3 & 0.28 & 6.2  & 0.32 \\
1024 & 6.6 & 0.27 & 18.4 & 0.92 & 7.3  & 0.65 \\
\bottomrule
\end{tabular}
\end{table}
\medskip

\subsection{Failure modes}
\label{sec:failure_modes}

\begin{figure}[!h]
  \centering
  \begin{minipage}[t]{0.49\linewidth}
    \centering
    \includegraphics[width=\linewidth]{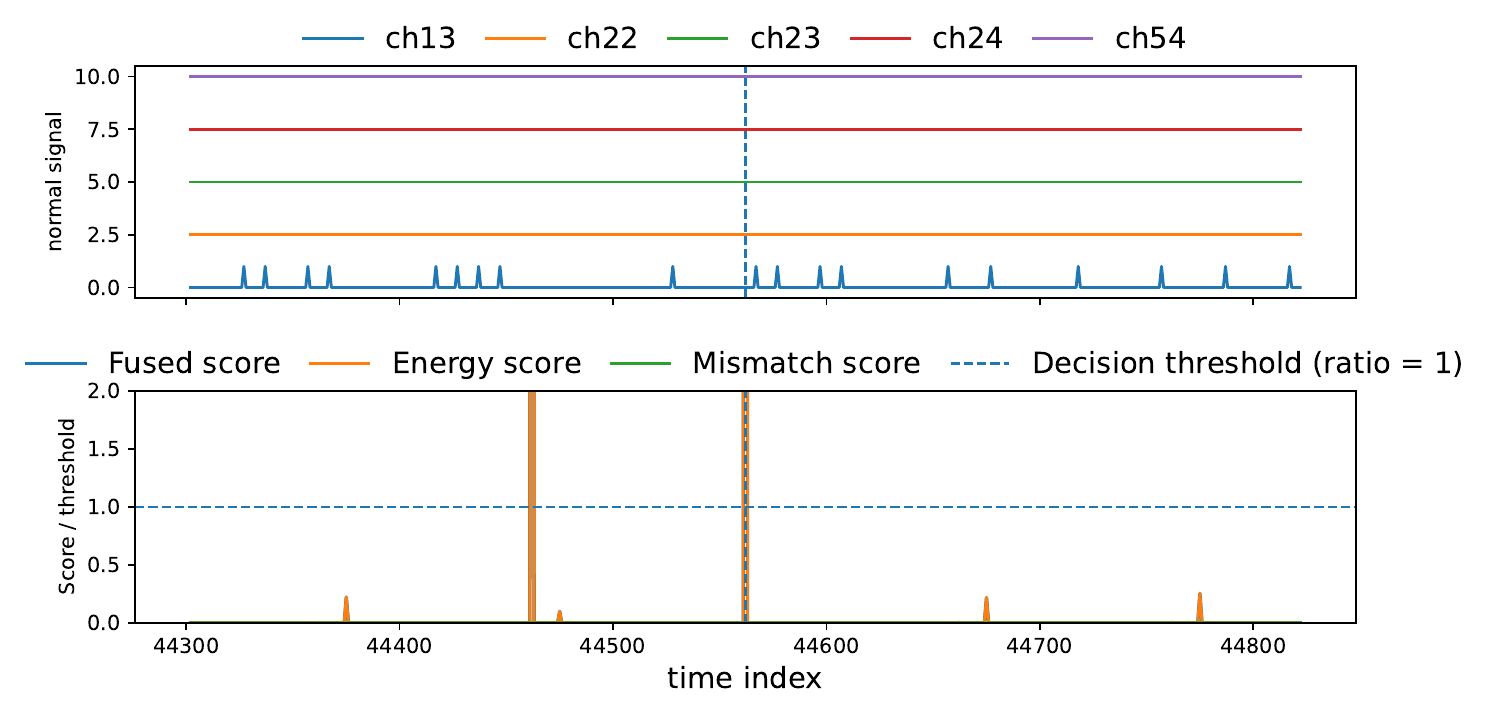}\\[-2pt]
    \footnotesize (a) MSL false positive.
  \end{minipage}\hfill
  \begin{minipage}[t]{0.49\linewidth}
    \centering
    \includegraphics[width=\linewidth]{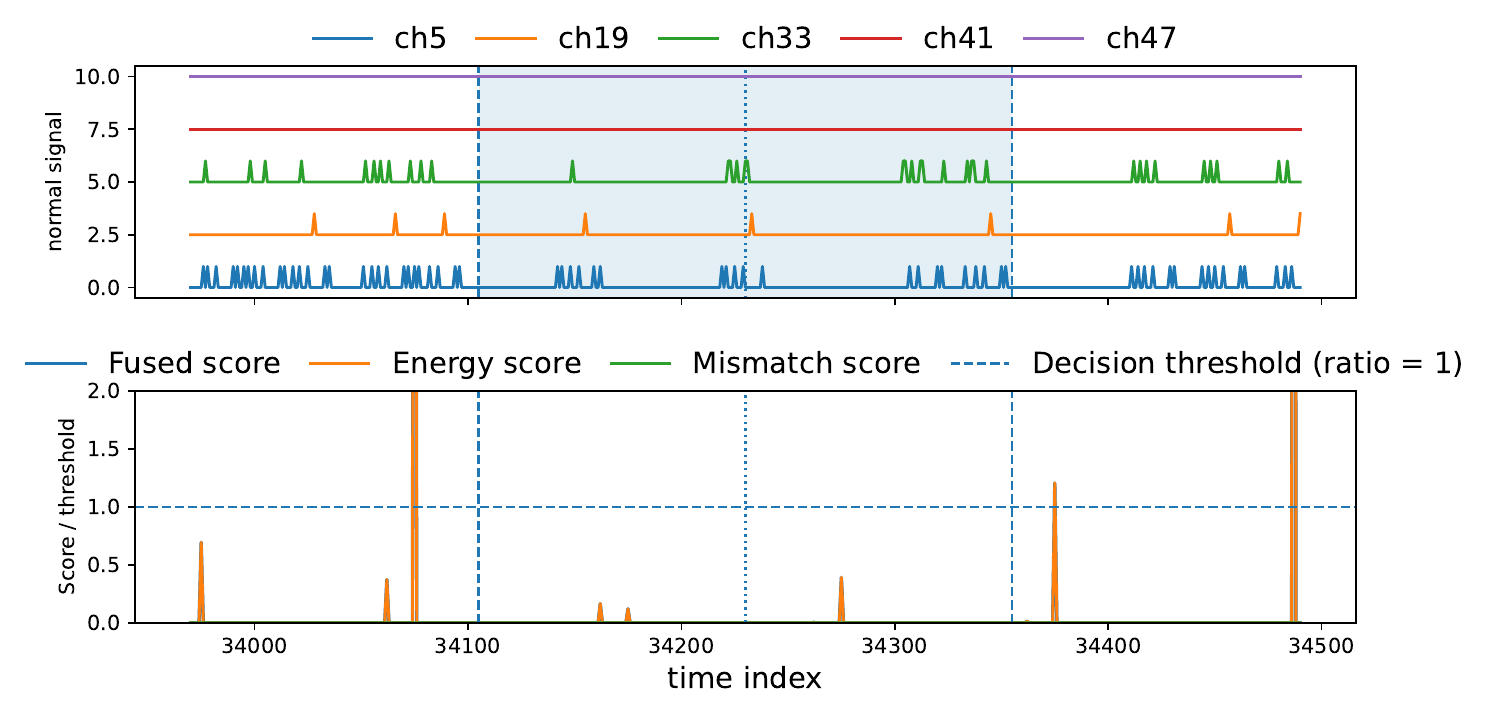}\\[-2pt]
    \footnotesize (b) MSL false negative.
  \end{minipage}
  \caption{\textbf{Representative MSL failure cases.} Top: selected channels (offset for readability). Bottom: Energy score, Mismatch score, and the fused score. The dashed line is the dataset decision threshold used in the main evaluation protocol.}
  \label{fig:failure_cases}
\end{figure}

\noindent We include two representative errors identified on MSL. Figure~\ref{fig:failure_cases}(a) shows a false positive. A short transient spike produces a strong increase in the Energy score, while the Mismatch score remains comparatively small. The fused score therefore crosses the dataset decision threshold, despite limited evidence of disrupted temporal coordination. Figure~\ref{fig:failure_cases}(b) shows a false negative. The event has limited impact on the Mismatch score and the Energy increase is not sufficient to push the fused score above the dataset decision threshold, leading to a missed detection.\medskip

\noindent These cases indicate that max fusion is most effective when either deviations persist sufficiently to raise Energy or coordination breaks in a way that raises the mismatch stream. Benign transients and weak or prior-consistent changes remain challenging, forming a clear operational boundary for the method.

\section{Conclusion}
\label{sec:conclusion}

\noindent Multivariate time series anomaly detection is a challenging task because anomalies do not simply show up as straightforward outliers. Instead, they arise from complex time-based relationships and subtle structural changes that might only be noticeable in how sensors coordinate. To address this, we introduced the Pi-Transformer, a prior-informed transformer model that combines the expressive capacity of attention with inductive biases to capture temporal invariants. The main idea is to treat reconstruction error as a sign of deviation while adjusting this evidence with a prior-alignment weight based on the agreement between data-driven series attention and a prior-informed attention. This design attempts to improve separability of anomaly evidence under benign variability. Concretely, the model produces an Energy stream and a mismatch stream; at inference we normalise both and form a fused score $f_i$ so the detector responds when either amplitude/shape evidence or a timing disruption is strong.\medskip

\noindent Testing on various benchmarking datasets showed that the Pi-Transformer achieves state-of-the-art or near state-of-the-art performance under the adopted point-adjust protocol. The model is particularly effective at detecting phase-breaking anomalies, where traditional methods often fail, while still accurately detecting amplitude-driven anomalies. Case studies demonstrated how the prior-alignment weights focus attention on stable areas near disruptions, how the Energy signal peaks in locations with high evidence but not complete alignment, and how the fused score rises at onsets where mismatch spikes even if Energy is briefly muted. Ablation studies confirmed the importance of the prior-informed attention pathway, showing that removing it caused notable drops in recall and overall F1 scores.\medskip

\noindent While the Pi-Transformer shows clear improvements, it also suggests paths for future development. The current use of stride-1 overlapping windows creates a slight timing offset between detected peaks and actual anomaly boundaries. This could be reduced through different score aggregation methods or by adjusting scores across windows. Accordingly, although the main benchmark results are strong under segment-level point-adjust evaluation, performance under stricter point-wise or range-aware metrics may be less favourable and should be examined in future work. Additionally, the model requires tuning of window length and inverse-temperature to balance accuracy against stability, which could benefit from adaptive or data-driven adjustments. Although the prior-informed attention is learned from data and regularised by smoothness and Hurst-related features, further investigation into application-specific priors may improve interpretability and robustness in non-stationary multivariate time series. In addition, scaling to higher-dimensional and longer-horizon monitoring settings is an important direction. Future work can investigate more efficient attention variants (e.g., sparse attention, patching, or hierarchical windowing) and implementation-level optimisations of the prior construction to reduce memory and runtime while preserving the dual-stream scoring. Where domain knowledge is available, the prior pathway could be augmented with stronger system-specific constraints (for example, known inter-channel coupling structure or causal priors) so the temporal reference encodes application-relevant invariants beyond generic timing relationships.

\noindent The Pi-Transformer model shows that integrating prior-informed structure into transformer models provides an effective way to detect anomalies in multivariate time series. By combining reconstruction evidence with prior-informed alignment via a fused score, it improves detection accuracy while providing interpretable signals that clarify the time dynamics of complex systems. This approach highlights the value of principled, prior-informed biases for developing robust and interpretable anomaly detectors in multivariate time series.

\bibliography{references}

\end{document}